\documentclass{article}
\PassOptionsToPackage{numbers, square}{natbib}
\usepackage[final]{neurips_2024}

\usepackage[utf8]{inputenc} 
\usepackage[T1]{fontenc}    
\usepackage{hyperref}       
\usepackage{url}            
\usepackage{booktabs}       
\usepackage{amsfonts}       
\usepackage{nicefrac}       
\usepackage{microtype}      
\usepackage{xcolor}         
\usepackage{graphicx}
\usepackage{wrapfig}
\usepackage{amsmath}
\usepackage{multirow}
\usepackage{subfig}
\usepackage{enumitem}
\usepackage[figuresright]{rotating}
\usepackage[normalem]{ulem}
\useunder{\uline}{\ul}{}

\title{Expanding Sparse Tuning for Low Memory Usage}

\author{
  Shufan Shen$^{1,2}$\quad
  Junshu Sun$^{1,2}$\quad
  Xiangyang Ji$^{3}$\quad
  Qingming Huang$^{1,2,4}$\quad
  Shuhui Wang$^{1,4}$\thanks{Corresponding author.}\and
  $^{1}$Key Lab of Intell. Info. Process., Inst. of Comput. Tech., CAS \\
  $^{2}$University of Chinese Academy of Sciences\quad
  $^{3}$Tsinghua University\quad
  $^{4}$Peng Cheng Laboratory\and
  \texttt{\{shenshufan22z, sunjunshu21s, wangshuhui\}@ict.ac.cn} \quad
  \texttt{xyji@tsinghua.edu.cn}\\
  \texttt{qmhuang@ucas.ac.cn}
}

\begin{document}

\maketitle

\begin{abstract}
  Parameter-efficient fine-tuning (PEFT) is an effective method for adapting pre-trained vision models to downstream tasks by tuning a small subset of parameters.
  Among PEFT methods, sparse tuning achieves superior performance by only adjusting the weights most relevant to downstream tasks, rather than densely tuning the whole weight matrix.
  However, this performance improvement has been accompanied by increases in memory usage, which stems from two factors, {\it i.e.}, the storage of the whole weight matrix as learnable parameters in the optimizer and the additional storage of tunable weight indexes.
  In this paper, we propose a method named SNELL~(\textbf{S}parse tuning with ker\textbf{NEL}ized \textbf{L}oRA) for sparse tuning with low memory usage. 
  To achieve low memory usage, SNELL decomposes the tunable matrix for sparsification into two learnable low-rank matrices, saving from the costly storage of the whole original matrix. A competition-based sparsification mechanism is further proposed to avoid the storage of tunable weight indexes.
  To maintain the effectiveness of sparse tuning with low-rank matrices, we extend the low-rank decomposition by applying nonlinear kernel functions to the whole-matrix merging. Consequently, we gain an increase in the rank of the merged matrix, enhancing the ability of SNELL in adapting the pre-trained models to downstream tasks.
  Extensive experiments on multiple downstream tasks show that SNELL achieves state-of-the-art performance with low memory usage, endowing PEFT with sparse tuning to large-scale models. 
  Codes are available at \href{https://github.com/ssfgunner/SNELL}{https://github.com/ssfgunner/SNELL}.
\end{abstract}
\section{Introduction}

Fine-tuning has become a predominant way for adapting large pre-trained models to downstream tasks with limited training samples~\cite{devlin2018bert,chen2020simple,he2020momentum,he2022masked}.
Nevertheless, fine-tuning all model parameters requires substantial memory usage and is susceptible to over-fitting, making it costly and infeasible for large-scale models~\cite{zhai2022scaling,bai2023sequential,dai2021coatnet}.
To address these limitations, parameter-efficient fine-tuning~(PEFT)~\cite{zhao2020masking,hu2021lora,zhang2022neural, jia2022visual, chen2022adaptformer, he2023sensitivity} has been proposed to tune a small subset of parameters while keeping other parameters frozen.
PEFT methods can be categorized into addition-based and reparameterization-based methods.
The former attaches additional parameters to a frozen pre-trained backbone, while the latter adjusts the original parameters in the pre-trained backbone.

\begin{figure}[tb]
  \centering
  \includegraphics[width=1.0\linewidth]{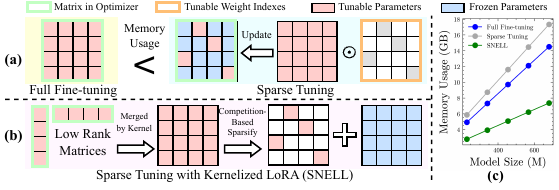}
  \vspace{-0.3cm}
  \caption{(a) The high memory usage of sparse tuning arises from taking the whole weight matrix as learnable parameters, in addition to the storage of the tunable weight indexes~(typically represented as a binary mask). (b) Our framework~(SNELL) only stores the learnable low-rank matrices in the optimizer. (c) Memory usage comparison on pre-trained models with different depths.}
  \label{fig: intro}
  \vspace{-0.3cm}
\end{figure}

Addition-based methods~\cite{tu2023visual, zhang2022neural, jia2022visual} have achieved remarkable performance on vision tasks. 
However, adopting additional parameters incurs extra computational costs during the inference process. 
In contrast, reparameterization-based methods~\cite{zaken2021bitfit, caelles2017one, hu2021lora} directly fine-tune the original parameters. These methods select specific parameters, involving reduced memory usage compared to full-parameter fine-tuning.
Based on the granularity of parameter selection, one primary approach focuses on specific parameter matrices. For example, Bitfit~\cite{zaken2021bitfit} only adjusts bias to reduce the volume of tunable parameters while Partial-\textit{k}~\cite{jia2022visual} fine-tunes the last few layers to avoid back-propagation through the entire pre-trained backbone. 
To further reduce memory usage, LoRA~\cite{hu2021lora} optimizes each selected weight matrix using two low-rank matrices to achieve memory-efficient fine-tuning.
Although sufficient in reducing memory usage, these methods usually gain inferior performance compared to addition-based methods~\cite{jia2022visual}.
Recently, SPT~\cite{he2023sensitivity} and GPS~\cite{DBLP:conf/cvpr/ZhangZGZSZZ24} found that combining existing PEFT methods with sparse tuning, which only adjusts the most task-related weights in a matrix, can achieve state-of-the-art performance on vision tasks. Concurrently, the effectiveness of sparse tuning has also been observed in NLP tasks~\cite{fu2023effectiveness}.
By focusing on individual weights in a matrix, sparse tuning allows for more precise adjustments, thus achieving good performance and mitigated over-fitting risks~\cite{fu2023effectiveness}.

However, the performance gained from sparse tuning has been accompanied by high memory usage, as Figure~\ref{fig: intro}(a) shows.
Although sparse tuning only updates part of weights in the pre-trained weight matrix, the whole matrix still needs to be stored as learnable parameters in the optimizer and computed for their corresponding gradients in practice.
Additionally, sparse tuning necessitates storing the tunable weight indexes, further aggravating the memory demands.
The above observation indicates that sparse tuning gains no advantage over full fine-tuning regarding memory usage, especially given the increasing parameter volumes in pre-trained models~\cite{zhai2022scaling, bai2023sequential}.
A sparse tuning method with low memory usage is urgently required for applications on large-scale pre-trained models.

In this paper, we propose a method that conducts \textbf{S}parse tuning with ker\textbf{NEL}ized \textbf{L}oRA~(SNELL) shown in Figure~\ref{fig: intro}(b). SNELL can adapt pre-trained models to downstream tasks with both low memory usage and strong performance.
To reduce memory usage, we decompose the tunable matrix for sparsification into low-rank learnable matrices to store fewer parameters in the optimizer and develop a competition-based method to avoid storing the tunable weight indexes.
To improve the performance on downstream tasks, we extend LoRA from a kernel perspective and merge low-rank matrices with nonlinear kernel functions to obtain matrices with higher ranks.

Specifically, SNELL updates the pre-trained weight matrix using a sparse low-rank adaptation matrix. This adaptation matrix is first merged with two low-rank learnable matrices and then sparsified toward effective fine-tuning. 
Compared to storing the whole adaptation matrix, storing low-rank matrices in the optimizer results in lower memory usage.
For the sparsification process, we propose a competition-based method inspired by the neuron competition phenomenon in neuroscience~\cite{sun2023stem}, avoiding the storage of the tunable weight indexes that incur additional memory usage.
The proposed method promotes competition among weights based on their absolute values. Most task-relevant weights are encouraged to have larger absolute values and survive during the fine-tuning process.
By setting a sparsity ratio as the hyperparameter and determining tunable weights based on their absolute values in an end-to-end manner, we can eliminate the storage of the tunable weight indexes.

In addition to low memory usage, the performance is also critical for model fine-tuning. 
However, directly merging two low-rank matrices through the inner product leads to the low-rank structure of the adaptation matrix, which narrows the optimization scope of tunable matrices and further limits the expressiveness of sparse tuning.
To overcome this bottleneck, we draw inspiration from DyN~\cite{pei2023dynamics} on weight matrix interpretation based on low-dimensional dynamical systems, and reformulate the merging process with nonlinear kernel functions that increase the rank of the merged adaptation matrix.
This new formulation enables a more expressive sparse tuning while maintaining a compact representation with low memory.

Extensive experiments are conducted on 24 downstream visual recognition tasks with both plain and hierarchical vision Transformer backbones under supervised and self-supervised pre-training. 
Results show that SNELL can gain the performance improvement of sparse tuning and the low memory usage of LoRA concurrently.
SNELL obtains the state-of-the-art performance on FGVC~(91.8\% \textit{vs.} 90.7\%) and VTAB-1k~(74.6\% \textit{vs.} 74.1\%) benchmark with LoRA-level memory usage.
Moreover, as Figure~\ref{fig: intro}(c) shows, the low memory-usage advantage of SNELL becomes increasingly apparent as the model size grows, enabling sparse tuning on larger models.

\section{Related Work}
\textbf{Parameter Sparsity.}
In early work, the parameter sparsity usually serves as an optimization objective in model pruning~\cite{han2016eie,molchanov2017variational}.
These pruning methods remove the weights from pre-trained models irrelevant to a specific task, without significantly degrading model performance. The relevance of individual weights can be estimated based on activations~\cite{hu2016network}, redundancy~\cite{srinivas2015data}, per-layer second derivatives~\cite{dong2017learning}, and energy efficiency~\cite{yang2017designing}. 
Except for the post-training pruning strategy, sparse networks~\cite{bellec2017deep,louizos2017learning,frankle2018lottery} directly introduce parameter sparsity into the training process, removing redundant weights more precisely~\cite{frankle2018lottery}.
Motivated by the advantage of parameter sparsity in model optimization, recent studies introduce sparsity to the fine-tuning of pre-trained models and achieve enhanced model performance on downstream tasks~\cite{ansell2021composable,han2016eie,xu2021raise}.
The parameter sparsity gives rise to a reduced number of trainable parameters and serves as a regularization constraint during fine-tuning~\cite{fu2023effectiveness}.
Among sparse tuning, pre-pruning methods adopt model pruning for fine-tuning. These methods sparsify the weight matrix~\cite{he2023sensitivity, DBLP:conf/cvpr/ZhangZGZSZZ24} or adapter~\cite{ansell2021composable} through pruning metrics~\cite{molchanov2019importance,frankle2018lottery} to identify learnable parameters for the fine-tuning process.
Other methods select trainable parameters during fine-tuning, including learnable mask~\cite{zhao2020masking} or diff vectors~\cite{guo2020parameter} with sparsity constraints.
However, the parameter sparsification methods need to store the indexes of tunable weights, which incurs additional memory usage.
For sparse tuning under low memory budget, our competition-based mechanism selects weights relevant to downstream tasks in a learnable manner without storing the tunable weight indexes. 

\textbf{Parameter-efficient Fine-tuning.}
Fine-tuning is the most predominant approach for adapting a pre-trained model to downstream tasks.
However, for large pre-trained models, fine-tuning all parameters is costly and prone to overfit downstream datasets.
To tackle these problems, parameter-efficient fine-tuning~(PEFT)~\cite{chen2022adaptformer, jia2022visual,zhang2022neural}, which tunes only a tiny portion of parameters, becomes a desirable choice.
Following the taxonomy of SPT~\cite{he2023sensitivity}, PEFT methods can be categorized into \textit{addition-based}~\cite{bapna2019simple,houlsby2019parameter,pfeiffer2020adapterfusion,sung2022vl,ding2021openprompt,ju2022prompting,liu2021p,zhang2022neural} and \textit{reparameterization-based}~\cite{zaken2021bitfit,caelles2017one,guo2020parameter,zhao2020masking,hu2021lora} methods. 

\textit{Addition-based} methods attach additional trainable parameters to a frozen pre-trained backbone.
Adapters~\cite{bapna2019simple,houlsby2019parameter,pfeiffer2020adapterfusion,sung2022vl,zhang2021tip} adopt a residual pathway and learn a bottleneck layer including two linear projections and a non-linear activation.
Prompt-tuning methods~\cite{ding2021openprompt,ju2022prompting,liu2021p,li2021prefix} add trainable parameters to the input and keep the entire pre-trained model unchanged during training.
Recent work~\cite{zhang2022neural} attempts to find the optimal configurations to combine multiple addition-based methods.
Despite of the popularity and effectiveness of \textit{addition-based} methods, the additional trainable parameters incur excess computational costs during the inference process~\cite{bapna2019simple,li2022cross}.

\textit{Reparametization-based} methods adjust the inherent parameters in the pre-trained backbone to avoid excess computational costs during inference.
Early work directly selects parameters with low memory usage for fine-tuning, such as the bias terms~\cite{zaken2021bitfit} and the final few layers of the pre-trained model~\cite{caelles2017one}.
To further reduce the memory usage of the selected matrices, LoRA~\cite{hu2021lora} optimizes low-rank matrices that can be reparameterized into the pre-trained weight matrices to reduce memory usage.
Exploring finer-grained parameter selection, some studies~\cite{guo2020parameter,zhao2020masking} propose sparse tuning, which involves selecting and tuning individual weights sparsely within the weight matrices.
Recently, SPT~\cite{he2023sensitivity} combines sparse tuning and LoRA in a hybrid framework that achieves state-of-the-art performances on visual PEFT tasks. 
SPT has revealed that optimizing the weights most relevant to the downstream task through sparse tuning can significantly enhance the performance, which is also supported by SAM~\cite{fu2023effectiveness} and GPS~\cite{DBLP:conf/cvpr/ZhangZGZSZZ24}.
However, existing sparse tuning framework still faces the challenge of high memory usage brought by sparse tuning.
Unlike existing methods, our SNELL inherits high performance and low memory usage concurrently by sparsifying the adaptation matrix merged with low-rank matrices through nonlinear kernels.
\section{Methodology}
We first introduce the definitions of sparse tuning~\cite{zhao2020masking, guo2020parameter, ansell2021composable}, LoRA~\cite{hu2021lora} and kernel trick~\cite{koutroumbas2008pattern}~(Section \ref{subsec: preliminaries}). 
Then we propose SNELL, a sparse tuning method including kernelized LoRA that enables high-performance tuning with low-rank learnable matrices~(Section \ref{subsec: k_lora}) and a competition-based mechanism that sparsifies weights without additional memory usage~(Section \ref{subsec: sparsification_mechanism}).

\subsection{Preliminaries}
\label{subsec: preliminaries}
\noindent\textbf{Sparse Tuning}.
Given a downstream training set $\mathcal{D}=\{x^{(n)}, y^{(n)}\}_{n=1}^N$, the objective of sparse tuning is to minimize the model's empirical risk on the downstream task, with the sparsity constraints on the volume of tunable weights in weight matrix $\mathbf{W}\in\mathbb{R}^{m\times n}$.
The sparsification is usually achieved through a binary mask $\mathbf{M}\in\{0,1\}^{m\times n}$. The objective function can be formulated as
\begin{equation}
    \min_{\mathbf{W}\odot\mathbf{M}} \frac{1}{N} \sum_{n=1}^N \mathcal{L}\left(f(x^{(n)} ; \mathbf{W}), y^{(n)}\right)
\label{eq: training_objective}
\end{equation}
where $f(\cdot; \cdot)$ is a parameterized function over the input (\textit{e.g.}, a neural network), $\mathcal{L}(\cdot, \cdot)$ is a loss function (\textit{e.g.}, cross-entropy), and $\odot$ denotes element-wise multiplication.
The binary mask $\mathbf{M}$ can be either a fixed hyperparameter, pre-computed with heuristics such as pre-pruning~\cite{he2023sensitivity}, or a learnable parameter obtained through end-to-end fine-tuning~\cite{zhao2020masking}.
All these methods require storing $\mathbf{M}$ to determine the tunable weights, which results in additional memory usage.
More importantly, the tunable parameters $\mathbf{W}\odot\mathbf{M}$ occupy the same amount of memory as the weight matrix $\mathbf{W}$ in practice. 
As a result, the memory usage of sparse tuning is even higher than that of full fine-tuning.

\noindent\textbf{LoRA}.
Given a pre-trained weight matrix $\mathbf{W}_0$, LoRA~\cite{hu2021lora} optimizes two low-rank matrices $\mathbf{B}\in\mathbb{R}^{m\times r}, \mathbf{A}\in\mathbb{R}^{n\times r}$ to reduce the memory usage during fine-tuning. 
The low-rank matrices $\mathbf{A}$ and $\mathbf{B}$ can be reparameterized into the pre-trained weight $\mathbf{W}_0$,
\begin{equation}
    \mathbf{W} = \mathbf{W}_0 + \Delta \mathbf{W} = \mathbf{W}_0 + \mathbf{BA}^\top
\end{equation}
With $r\ll \min(m,n)$, LoRA can achieve high training efficiency and low memory usage by only optimizing the smaller low-rank matrices.

\noindent\textbf{Kernel Trick}~\cite{koutroumbas2008pattern}.
In many machine learning tasks, mapping the vectors into higher dimensions is frequently used to achieve linear separability~\cite{suykens1999chaos}.
However, the explicit mapping process incurs significant computational costs. To address this problem, the kernel trick is proposed to efficiently model data relationships in high-dimensional spaces, without the need to explicitly formulate the space. According to Mercer's theorem~\cite{berlinet2011reproducing}, a kernel function $\kappa: \mathbb{R}^r\times \mathbb{R}^r \rightarrow \mathbb{R}$ can express an inner product in some space as $\kappa(\mathbf{x}, \mathbf{x'}) = \phi(\mathbf{x})^\top \phi(\mathbf{x'})$, if and only if $\kappa$ is positive semi-definite~(Appendix~\ref{sec: app_kernel}). $\mathbf{x}, \mathbf{x'} \in \mathbb{R}^r$, and $\phi: \mathbb{R}^r \rightarrow \mathbb{R}^d$ is an implicit feature map.
By selecting an appropriate kernel function $\kappa$, we can obtain the inner product of two vectors in higher-dimensional space $\mathbb{R}^d~(d\ge r)$ without explicitly formulating the feature map $\phi$.

\begin{figure}[tb]
  \centering
  \includegraphics[width=1.0\linewidth]{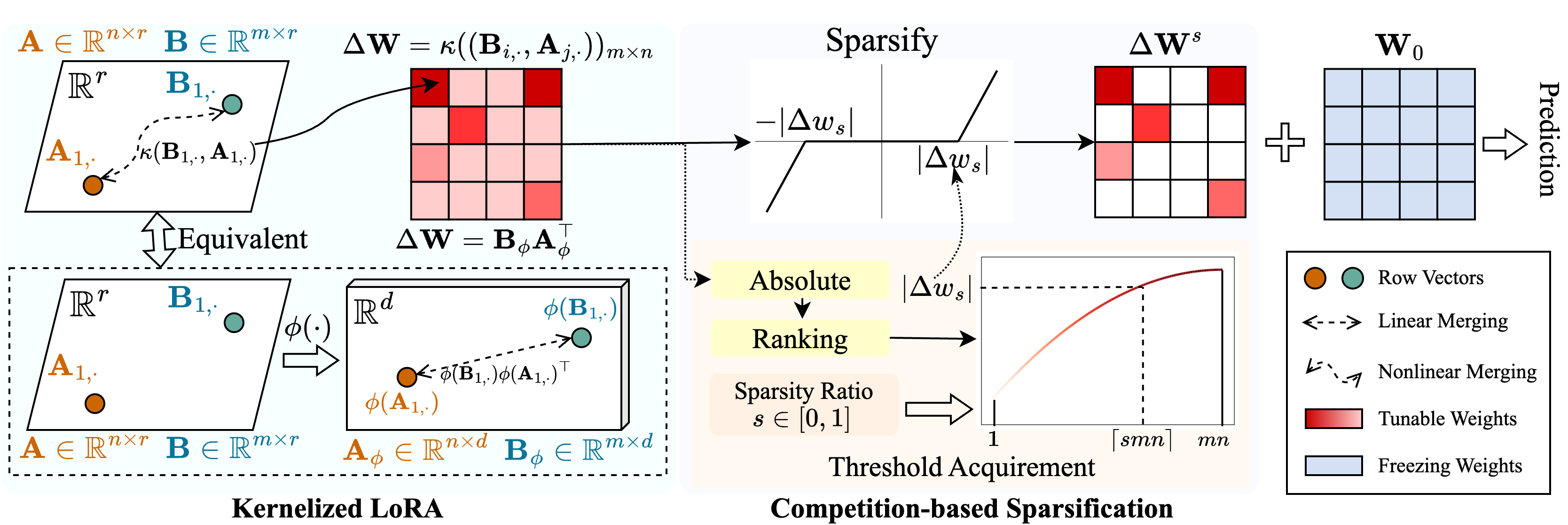}
  \vspace{-0.3cm}
  \caption{Overview of our SNELL strategy. Given two learnable low-rank matrices, we merge them using a non-linear kernel function~(\textit{left}). This merging process is equivalent to mapping the matrices to higher-rank matrices and then performing matrix multiplication. Then we sparsified this merged adaptation matrix using a competition-based sparsification mechanism~(\textit{right}). This mechanism zeros out weights with small absolute values based on the specified percentage of $s$. }
  \label{fig: framework}
  \vspace{-0.4cm}
\end{figure}

\subsection{Kernelized LoRA}
\label{subsec: k_lora}
We leverage LoRA to reduce the memory usage of sparse tuning in light of its low memory usage.
An intuitive solution is to sparsify the adaptation matrix $\Delta \mathbf{W}$ composed of the two low-rank matrices.
However, the low-rank property of $\Delta \mathbf{W}$ can lead to the performance degradation of sparse tuning.
For the original sparse tuning, the weight matrix $\mathbf{W}$ is free of the rank constraint, and weights are independent of each other. 
Therefore, we can independently select and optimize weights most relevant to the downstream task.
For sparse tuning with LoRA, the adaptation matrix $\Delta \mathbf{W}$ with rank $r$ is constrained in $\mathbb{R}^{r\times (m+n-r)}$, a subspace of $\mathbb{R}^{m\times n}$.
When $r \ll \min(m, n)$, the weight optimization scope of sparse tuning contracts, hindering its performance on downstream tasks.

To achieve sparse tuning with both strong performance and low memory usage, we propose to construct a high-rank matrix using low-rank matrices.
Inspired by DyN~\cite{pei2023dynamics} that fits a high-rank matrix using the distance matrix of a low-dimension dynamical system, we extend the distance function to general kernel functions and investigate LoRA in the kernel perspective.
Given two vectors $\boldsymbol{x}, \boldsymbol{x'}\in\mathbb{R}^r$, the kernel function $\kappa(\boldsymbol{x}, \boldsymbol{x'})$ can be formulated as an inner product $\phi(\boldsymbol{x})^\top\phi(\boldsymbol{x'})$ with an implicit feature map $\phi: \mathbb{R}^r \rightarrow \mathbb{R}^d$.
The merging process of LoRA can be seen as applying linear kernel function $\kappa_l(\cdot,\cdot)$ on the rows of the learnable parameters $\mathbf{A}$ and $\mathbf{B}$,
\begin{equation}
    \Delta \mathbf{W}_{ij} = \kappa_l(\mathbf{A}_{i,\cdot}, \mathbf{B}_{j,\cdot}) = \phi_l(\mathbf{B}_{j,\cdot})\phi_l(\mathbf{A}_{i,\cdot})^\top=\mathbf{B}_{j,\cdot}\mathbf{A}_{i,\cdot}^\top,
\end{equation}
where $\mathbf{A}_{i,\cdot}, \mathbf{B}_{j,\cdot}\in\mathbb{R}^r$, $\phi_l: \mathbb{R}^r \rightarrow \mathbb{R}^r$ denotes the identity mapping.
By replacing $\kappa_l(\cdot,\cdot)$ with more complex non-linear kernel functions, we can approximate relations in higher-dimensional spaces $\mathbb{R}^d$ and obtain matrices with rank larger than $r$.
The merged adaptation matrix in SNELL can be represented by 
\begin{equation}
    \Delta \mathbf{W} = (\kappa(\mathbf{A}_{i,\cdot}, \mathbf{B}_{j,\cdot}))_{m\times n} =  [\phi(\mathbf{B}_{1,\cdot})^\top,...,\phi(\mathbf{B}_{n,\cdot})^\top]^\top[\phi(\mathbf{A}_{1,\cdot})^\top,...,\phi(\mathbf{A}_{n,\cdot})^\top] = \mathbf{B}_\phi \mathbf{A}_\phi^\top.
\end{equation}
Note that in practice, explicit computation of $\mathbf{A}_\phi\in\mathbb{R}^{n\times d}$ and $\mathbf{B}_\phi\in\mathbb{R}^{m\times d}$ is unnecessary. $\Delta \mathbf{W}$ can be directly derived based on $\mathbf{A}$ and $\mathbf{B}$ with the kernel function $\kappa$.
By extending LoRA in a kernel perspective, SNELL can build high-rank adaptation matrices based on low-rank learnable matrices, empowering strong sparse tuning with low memory usage. We utilize the piecewise linear kernel introduced in Appendix~\ref{sec: app_kernel} without a specific statement.

\subsection{Competition-based Sparsification Mechanism}
\label{subsec: sparsification_mechanism}
Existing methods store tunable weight indexes $\mathbf{M}\in\{0,1\}^{m\times n}$ for sparsifying the update of the weight matrix $\mathbf{W}\in\mathbb{R}^{m\times n}$. The storage of $\mathbf{M}$ leads to additional memory usage.
Inspired by the neuron competition phenomenon in neuroscience~\cite{sun2023stem}, we design a competition-based parameter sparsification mechanism to avoid this additional storage. 
Instead of determining the learnable weights in the optimization process based on $\mathbf{M}$, our objective is to encourage the weights to compete based on their contributions to performance improvement. Weights with stronger contributions survive in the sparsification while the remaining low-contributed weights are zeroed out.
The weight contribution is reflected in their absolute values during the end-to-end optimization. 
During optimization, weights contributing more to the loss reduction are encouraged to have more significant values, while weights contributing less approach zero.
By retaining higher importance to significant weights and zeroing out the less impactful weights, we can achieve end-to-end tunable parameter selection by solely relying on the absolute values of weights, avoiding the storage of $\mathbf{M}$.

Specifically, given a merged adaptation matrix $\Delta \mathbf{W}$ and a sparsity ratio $s\in[0,1]$, we sparsify weights with a soft-threshold function. 
To induce weight competition during end-to-end fine-tuning, we propose a dynamic threshold $\Delta w_s$, \textit{i.e.}, the weight having the $\lceil smn \rceil$-th smallest absolute value in $\Delta \mathbf{W}$. This threshold ensures that only a fixed proportion~($s\times 100\%$) of weights remain non-zero. Therefore, the weights have to compete with each other to be selected instead of just having a larger absolute value than a fixed threshold.
\begin{equation}
    \Delta \mathbf{W}_{ij}^s = \Delta \mathbf{W}_{ij}\max(\vert\Delta \mathbf{W}_{ij}\vert-\vert\Delta w_s\vert, 0),
\end{equation}
where $\Delta \mathbf{W}^s=(\Delta \mathbf{W}_{ij}^{s})_{m\times n}$ denotes the sparse matrix with sparsity ratio $s$.
In practice, the sparsity ratio $s$ is manually determined regarding specific downstream tasks.
Given a sparsity ratio $s$, the training objective in Equation~\ref{eq: training_objective} can be reformulated as
\begin{equation}
    \min_{\mathbf{A}, \mathbf{B}} \frac{1}{N} \sum_{n=1}^N \mathcal{L}\left(f(x^{(n)} ; \mathbf{W}_0+\Delta \mathbf{W}^s), y^{(n)}\right).
\end{equation}
This objective encourages weights that are most relevant to the downstream task to gain more significant values for survival.
Adjusting the sparsity ratio $s$ allows us to control the sparsification process precisely and identify the optimal number of tunable parameters for different tasks.

\section{Experiments}

\subsection{Experimental Setup}
\label{sec: implementation}
\noindent\textbf{Datasets and Metrics.}
We evaluate our methods on 24 downstream tasks categorized into two groups following SPT~\cite{he2023sensitivity}.
\textit{(i) FGVC}~\cite{jia2022visual} is a benchmark for fine-grained image classification. This benchmark includes 5 downstream tasks, which are CUB-200-2011~\cite{wah2011caltech}, NABirds~\cite{van2015building}, Oxford Flowers~\cite{nilsback2008automated}, Stanford Dogs~\cite{gebru2017fine} and Stanford Cars~\cite{dataset2011novel}. 
We follow the validation splits in \cite{he2023sensitivity} if the official validation set is unavailable.
\textit{(ii) VTAB-1k}~\cite{zhai2019large} is a large-scale transfer learning benchmark consisting of 19 visual classification tasks. VTAB-1k can be further divided into three groups, {\it i.e.}, natural tasks with natural images, specialized tasks with images captured by specialized equipment, and structured tasks with images mostly generated from synthetic environments. 
We use top-1 accuracy averaged within each group as our main metric following \cite{he2023sensitivity}.

\noindent\textbf{Pre-trained Backbones.}
We conduct experiments on the plain vision Transformer backbone ViT-B/16~\cite{dosovitskiy2020image} that is pre-trained on ImageNet~\cite{russakovsky2015imagenet} with different pre-training strategies following~\cite{he2023sensitivity}, including supervised pre-training and self-supervised pre-training with MAE~\cite{he2022masked} and MoCo v3~\cite{chen2021empirical}. We also conduct experiments on the representative hierarchical vision Transformer backbone Swin-B~\cite{liu2021swin} and CNN backbone ConvNeXt-Base~\cite{liu2022convnet} under supervised pre-training.
In addition, we fine-tune the supervised pre-trained large-scale models~(ViT-L/16~\cite{dosovitskiy2020image}, ViT-H/14~\cite{dosovitskiy2020image}) on VTAB-1k to demonstrate the memory-efficiency and high-performance of SNELL.

\noindent\textbf{Competitors.}
We compare our methods with addition-based methods including MLP-\textit{k}, VPT-Shallow~\cite{jia2022visual}, VPT-Deep~\cite{jia2022visual}, Adapter-$r$~\cite{houlsby2019parameter}, and SPT-Adapter~\cite{he2023sensitivity}.
For reparameterization-based methods, we compare with Linear, Partial-1, Bias~\cite{zaken2021bitfit}, LoRA-$r$~\cite{hu2021lora}, SSF~\cite{lian2022scaling}, and SPT-LoRA~\cite{he2023sensitivity}. 
Here $r$ represents the number of bottleneck dimensions in Adapter-$r$ and the value of rank in LoRA-$r$ and our proposed SNELL-$r$. 
Details of the competitors are presented in Appendix~\ref{sec: app_contenders}. We also provide additional comparisons with other approaches~\cite{DBLP:conf/cvpr/ZhangZGZSZZ24, tu2023visual} in Appendix~\ref{sec: app_more_compare}.

\noindent\textbf{Implementation Details.}
Following SPT~\cite{he2023sensitivity}, we use the AdamW optimizer~\cite{loshchilov2018fixing} with cosine learning rate decay.
The batch size, learning rate, and weight decay are $32$, $1e-3$, and $1e-4$, respectively. We also follow SPT~\cite{he2023sensitivity} to implement the standard data augmentation pipeline for VTAB-1K and follow SSF~\cite{lian2022scaling} for FGVC as well. 
SNELL is applied on the pre-trained weight matrix of all linear layers. For each task, we fine-tune the model with different sparsity ratios $s$ to search the optimal volume of tunable parameters for this task. 
Without specific stating, we adopt the piecewise linear kernel~(introduced in Appendix~\ref{sec: app_kernel}) as the kernel function for SNELL. Ablation studies on different kernel functions are presented in Figure~\ref{fig: kernel_func}. 

\subsection{Performance on Downstream Tasks}

\textbf{Performance on Different Benchmarks.}
Experiments on FGVC and VTAB-1k benchmarks indicate that SNELL achieves the best performance with supervised pre-trained ViT-B/16 backbone as shown in Table~\ref{tab: fgvc_vtab1k}.
SNELL gains large performance improvements over LoRA variants, {\it e.g.}, SNELL-8 surpasses LoRA-8 significantly by 5.5\% in terms of mean accuracy on the FGVC benchmark. Moreover, SNELL outperforms the state-of-the-art method SPT-LoRA by a clear margin of 0.5\% in terms of mean top-1 accuracy on the VTAB-1k benchmark.
This stems from the fact that SPT-LoRA only performs sparse tuning on a portion of the weight matrices while employing LoRA for the remaining part. In contrast, the low memory property of SNELL empowers sparse tuning on all the weight matrices, allowing for more precise adjustments and giving rise to superior performance.
\begin{table}[]
\centering
\caption{Top-1 accuracy (\%) on FGVC and VTAB-1k benchmarks using ViT-B/16 pre-trained on ImageNet-21k supervisedly. The best result is in \textbf{bold}, and the second-best result is {\ul underlined}.}
\setlength\tabcolsep{3.5pt}
\begin{footnotesize}
\scalebox{0.8}{\begin{tabular}{@{}ccccccccccc@{}}
\toprule
\multicolumn{1}{l|}{\multirow{2}{*}{Method}} & \multicolumn{6}{c|}{FGVC}                                                                           & \multicolumn{4}{c}{VTAB-1k}                                   \\ \cmidrule(l){2-11} 
\multicolumn{1}{c|}{}                        & CUB-200 & NABirds & \begin{tabular}[c]{@{}c@{}}Oxford\\ Flowers\end{tabular} & \begin{tabular}[c]{@{}c@{}}Stanford\\ Dogs\end{tabular} & \begin{tabular}[c]{@{}c@{}}Stanford\\ Cars\end{tabular} & \multicolumn{1}{c|}{Mean Acc.} & Natural       & Specialized   & Structured    & Mean Acc.     \\ \midrule
\multicolumn{1}{l|}{Full}                    & 87.3    & 82.7    & 98.8           & 89.4          & 84.5          & \multicolumn{1}{c|}{88.5}      & 75.9          & 83.4          & 47.6          & 69.0          \\ \midrule
\multicolumn{11}{c}{Additional-based methods}                                                                                                                                                                      \\ \midrule
\multicolumn{1}{l|}{MLP-3~\cite{jia2022visual}}                   & 85.1    & 77.3    & 97.9           & 84.9          & 53.8          & \multicolumn{1}{c|}{79.8}      & 67.8          & 72.8          & 30.6          & 57.1          \\
\multicolumn{1}{l|}{VPT-Shallow~\cite{jia2022visual}}             & 86.7    & 78.8    & 98.4           & 90.7          & 68.7          & \multicolumn{1}{c|}{84.6}      & 76.8          & 79.7          & 47.0          & 67.8          \\
\multicolumn{1}{l|}{VPT-Deep~\cite{jia2022visual}}                & 88.5    & 84.2    & 99.0           & 90.2          & 83.6          & \multicolumn{1}{c|}{89.1}      & 78.5          & 82.4          & 55.0          & 72.0          \\
\multicolumn{1}{l|}{Adapter-8~\cite{houlsby2019parameter}}               & 87.3    & 84.3    & 98.4           & 88.8          & 68.4          & \multicolumn{1}{c|}{85.5}      & 79.0          & 84.1          & 58.5          & 73.9          \\
\multicolumn{1}{l|}{Adapter-32~\cite{houlsby2019parameter}}              & 87.2    & 84.3    & 98.5           & 89.6          & 68.4          & \multicolumn{1}{c|}{85.6}      & 79.6          & 84.0          & 58.3          & 74.0          \\
\multicolumn{1}{l|}{SPT-Adapter~\cite{he2023sensitivity}}             & 89.1    & 83.3    & 99.2           & 91.1          & 86.2          & \multicolumn{1}{c|}{89.8}      & 82.0          & 85.8          & 61.4          & 76.4\\
\multicolumn{1}{l|}{MoSA~\cite{zhang2023mosa}}             & 89.3    & 85.7    & 99.2           & 91.9          & 83.4          & \multicolumn{1}{c|}{89.9}      & 79.9          & 84.0          & 50.3         & 71.4\\ \midrule
\multicolumn{11}{c}{Reparameter-based methods}                                                                                                                                                                     \\ \midrule
\multicolumn{1}{l|}{Linear~\cite{jia2022visual}}                  & 85.3    & 75.9    & 97.9           & 86.2          & 51.3          & \multicolumn{1}{c|}{79.3}      & 68.9          & 77.2          & 26.8          & 57.6          \\
\multicolumn{1}{l|}{Partial-1~\cite{jia2022visual}}               & 85.6    & 77.8    & 98.2           & 85.5          & 66.2          & \multicolumn{1}{c|}{82.6}      & 69.4          & 78.5          & 34.2          & 60.7          \\
\multicolumn{1}{l|}{Bias~\cite{zaken2021bitfit}}                    & 88.4    & 84.2    & 98.8           & 91.2          & 79.4          & \multicolumn{1}{c|}{88.4}      & 73.3          & 78.3          & 44.1          & 65.2          \\
\multicolumn{1}{l|}{LoRA-8~\cite{hu2021lora}}                  & 84.9    & 79.0    & 98.1           & 88.1          & 79.8          & \multicolumn{1}{c|}{86.0}      & 79.5          & 84.6          & 60.5          & 74.9          \\
\multicolumn{1}{l|}{LoRA-16~\cite{hu2021lora}}                 & 85.6    & 79.8    & 98.9           & 87.6          & 72.0          & \multicolumn{1}{c|}{84.8}      & 79.8          & 84.9          & 60.2          & 75.0          \\
\multicolumn{1}{l|}{SPT-LoRA~\cite{he2023sensitivity}}                & 88.6    & 83.4    & {\ul 99.5}           & 91.4          & 87.3          & \multicolumn{1}{c|}{90.1}      & 81.9          & 85.9    & 61.3          & 76.4          \\
\multicolumn{1}{l|}{SSF~\cite{lian2022scaling}}                & 89.5    & 85.7    & \textbf{99.6}           & 89.6          & 89.2          & \multicolumn{1}{c|}{90.7}      & 81.6          & {\bf 86.6}    & 59.0          & 75.7          \\ \midrule
\multicolumn{1}{l|}{SNELL-8~(\textbf{ours})}                 & {\ul 89.6}    & {\ul 86.8}    & {99.3}           & {\ul 92.1}          & 89.9          & \multicolumn{1}{c|}{91.5}      & 82.0          & 85.7          & 61.6          & 76.4          \\
\multicolumn{1}{l|}{SNELL-16~(\textbf{ours})}                & {\bf 89.9}    & {\bf 87.0}    & 99.3           & {\bf 92.2}          & {\ul 90.3}          & \multicolumn{1}{c|}{{\ul 91.7}}      & {\ul 82.4}    & {\ul 86.1} & {\ul 61.7}    & {\ul 76.7}    \\
\multicolumn{1}{l|}{SNELL-32~(\textbf{ours})}                & {\bf 89.9}    & {\bf 87.0}    & {\ul 99.4}           & 92.0          & \textbf{90.5}          & \multicolumn{1}{c|}{\textbf{91.8}}          & \textbf{82.7} & {\ul 86.1} & \textbf{61.8} & \textbf{76.9} \\ \bottomrule
\end{tabular}}
\end{footnotesize}
\label{tab: fgvc_vtab1k}
\vspace{-0.7cm}
\end{table}
\begin{table}[]
\centering
\footnotesize
\setlength\tabcolsep{8.2pt}
\caption{Top-1 accuracy (\%) on VTAB-1k benchmarks using ViT-B/16 backbone pre-trained on ImageNet using MAE and MoCo v3 strategies. The best result is in \textbf{bold}.}
\scalebox{0.8}{
\begin{tabular}{@{}l|cccc|cccc@{}}
\toprule
\multirow{2}{*}{Methods} & \multicolumn{4}{c|}{VTAB-1k MAE}                & \multicolumn{4}{c}{VTAB-1k MoCo v3}            \\ \cmidrule(l){2-9} 
                         & Natural & Specialized & Structured & Mean Acc. & Natural & Specialized & Structured & Mean Acc. \\ \midrule
Full                     & 59.3    & 79.7        & 53.8       & 64.3      & 72.0    & 84.7        & 42.0       & 69.6      \\ \midrule
\multicolumn{9}{c}{Additional-based methods}                                                                               \\ \midrule
Adapter-8~\cite{houlsby2019parameter}                & 57.2    & 78.4        & 54.7       & 63.4      & 27.6    & 70.9        & 48.4       & 49.0      \\
Adapter-32~\cite{houlsby2019parameter}               & 55.3    & 78.8        & 53.3       & 62.5      & 74.2    & 82.7        & 47.7       & 68.2      \\
VPT-Shallow~\cite{jia2022visual}              & 40.0    & 69.7        & 27.5       & 45.7      & 67.3    & 82.3        & 37.6       & 62.4      \\
VPT-Deep~\cite{jia2022visual}                 & 36.0    & 60.6        & 26.6       & 41.1      & 70.3    & 83.0        & 42.4       & 65.2      \\
SPT-Adapter~\cite{he2023sensitivity}              & 65.6    & 82.7        & 60.7       & 69.7      & 76.6    & 85.0        & 61.7       & 74.4      \\ \midrule
\multicolumn{9}{c}{Reparameterization-based methods}                                                                       \\ \midrule
Linear~\cite{jia2022visual}                   & 18.9    & 52.7        & 23.7       & 32.1      & 67.5    & 81.1        & 30.3       & 59.6      \\
Partial-1~\cite{jia2022visual}                & 58.4    & 78.3        & 47.6       & 61.5      & 72.3    & 84.6        & 47.9       & 68.3      \\
Bias~\cite{zaken2021bitfit}                     & 54.6    & 75.7        & 47.7       & 59.3      & 72.9    & 81.1        & 53.4       & 69.2      \\
LoRA-8~\cite{hu2021lora}                   & 57.5    & 77.7        & 57.7       & 64.3      & 21.2    & 66.7        & 45.1       & 44.3      \\
LoRA-16~\cite{hu2021lora}                  & 57.3    & 77.1        & 59.9       & 64.8      & 16.0    & 64.0        & 48.7       & 42.9      \\
SPT-LoRA~\cite{he2023sensitivity}                 & 65.4    & 82.4        & 61.5       & 69.8      & 76.5    & \textbf{86.0}        & 63.6       & 75.3      \\ \midrule
SNELL-8~(\textbf{ours})           & \textbf{68.3}       & \textbf{83.8}           & \textbf{63.5}          & \textbf{71.8}         & \textbf{76.8}       & \textbf{86.0}            & \textbf{63.7}          & \textbf{75.5}         \\ \bottomrule
\end{tabular}}
\label{tab: mae_moco}
\vspace{-0.7cm}
\end{table}

\noindent\textbf{Performance on Different Pre-training Strategies.}
Experimental results on models pre-trained using different strategies are presented in Table~\ref{tab: mae_moco}. SNELL outperforms the state-of-the-art performances on models pre-trained with MAE~(71.8\% \textit{vs.} 69.8\%) and MoCo v3~(75.5\% \textit{vs.} 75.3\%).
Furthermore, SNELL consistently outperforms other PEFT methods on every group of downstream datasets.
This demonstrates the general effectiveness of SNELL under different pre-training strategies.

\noindent\textbf{Performance on Different Architectures.}
Following VPT~\cite{jia2022visual} and SPT~\cite{he2023sensitivity}, we apply SNELL to the hierarchal vision transformer Swin-B and the CNN architecture ConvNeXt-Base. Experimental results are shown in Table~\ref{tab: swin_convnext}.
Results on Swin-B demonstrate that SNELL-8 outperforms existing reparameterization-based PEFT methods by 0.3\% and achieves comparable performance to the state-of-the-art addition-based method SPT-Adapter.
For ConvNeXt-Base, SNELL achieves a performance improvement of 0.4\% compared to the best-reported result.
These results obtained on different architectures further validate the versatility and effectiveness of our SNELL approach.

\begin{table}[]
\centering
\footnotesize
\setlength\tabcolsep{8pt}
\caption{Comparisons on VTAB-1k benchmark with supervised pre-trained Swin-B and ConvNeXt-B. Top-1 accuracy (\%) is reported. The best result is in \textbf{bold}.}
\scalebox{0.8}{
\begin{tabular}{@{}ccccccccc@{}}
\toprule
\multicolumn{1}{l|}{\multirow{2}{*}{Methods}} & \multicolumn{4}{c|}{VTAB-1k Swin-B}                                                & \multicolumn{4}{c}{VTAB-1k ConvNeXt-B}                          \\ \cmidrule(l){2-9} 
\multicolumn{1}{c|}{}                         & Natural       & Specialized   & Structured    & \multicolumn{1}{c|}{Mean Acc.}     & Natural       & Specialized   & Structured    & Mean Acc.     \\ \midrule
\multicolumn{1}{l|}{Full}                     & 79.1          & 86.2          & 59.7          & \multicolumn{1}{c|}{75.0}          & 78.0          & 83.7          & 60.4          & 74.0          \\ \midrule
\multicolumn{9}{c}{Additional-based methods}                                                                                                                                                       \\ \midrule
\multicolumn{1}{l|}{MLP-3~\cite{jia2022visual}}                    & 73.6          & 75.2          & 35.7          & \multicolumn{1}{c|}{61.5}          & 73.8          & 81.4          & 35.7          & 63.6          \\
\multicolumn{1}{l|}{VPT-Deep~\cite{jia2022visual}}                 & 76.8          & 84.5          & 53.4          & \multicolumn{1}{c|}{71.6}          & 78.5          & 83.0          & 44.6          & 68.7          \\
\multicolumn{1}{l|}{Adapter-8~\cite{houlsby2019parameter}}                & 81.7          & 87.3          & 61.2          & \multicolumn{1}{c|}{76.7}          & 83.1          & 84.9          & 64.6          & 77.5          \\
\multicolumn{1}{l|}{SPT-Adapter~\cite{he2023sensitivity}}              & 83.0          & 87.3          & \textbf{62.1} & \multicolumn{1}{c|}{\textbf{77.5}} & 83.7          & 86.2          & 65.3          & 78.4          \\ \midrule
\multicolumn{9}{c}{Reparameterization-based methods}                                                                                                                                               \\ \midrule
\multicolumn{1}{l|}{Linear~\cite{jia2022visual}}                   & 73.5          & 80.8          & 33.5          & \multicolumn{1}{c|}{62.6}          & 74.5          & 81.5          & 34.8          & 63.6          \\
\multicolumn{1}{l|}{Partial-1~\cite{jia2022visual}}                & 73.1          & 81.7          & 35.0          & \multicolumn{1}{c|}{63.3}          & 73.8          & 81.6          & 39.6          & 65.0          \\
\multicolumn{1}{l|}{LoRA-8~\cite{hu2021lora}}                   & 81.7          & 87.2          & 60.1          & \multicolumn{1}{c|}{76.3}          & 82.2          & 84.7          & 64.1          & 77.0          \\
\multicolumn{1}{l|}{SPT-LoRA~\cite{he2023sensitivity}}                 & 83.1          & 87.4          & 60.4          & \multicolumn{1}{c|}{77.2}          & 83.4          & 86.7          & \textbf{65.9} & 78.7          \\ \midrule
\multicolumn{1}{l|}{SNELL-8~(\textbf{ours})}                  & \textbf{83.3} & \textbf{87.7} & 61.4          & \multicolumn{1}{c|}{\textbf{77.5}} & \textbf{84.5} & \textbf{87.4} & 65.6          & \textbf{79.1} \\ \bottomrule
\end{tabular}}
\label{tab: swin_convnext}
\vspace{-0.4cm}
\end{table}

\begin{figure}
  \centering
  \includegraphics[width=1.0\linewidth]{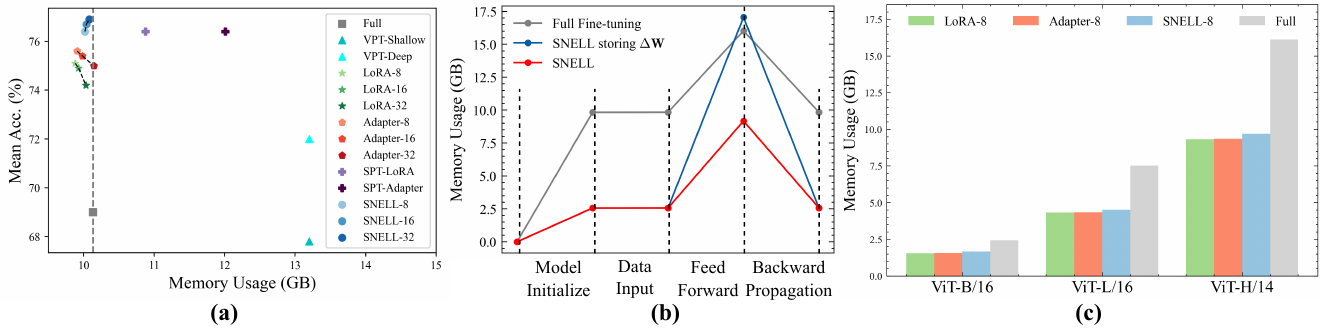}
  \vspace{-0.6cm}
  \caption{(a) Accuracy \textit{vs.} memory usage~(batchsize=64) with supervised pre-trained ViT-B/16 on VTAB-1k. 
  (b) Memory usage evolutions of full fine-tuning, SNELL, and SNELL storing the merged adaptation matrix~(SNELL storing $\Delta \mathbf{W}$) on ViT-H/14 during fine-tuning~(batchsize=8).
  (c) Model parameter volumes \textit{vs.} memory usage~(batchsize=8). As the model gets larger, SNELL's advantage of low memory usage over full fine-tuning becomes more obvious.}
  \label{fig: memory_analysis}
  \vspace{-0.6cm}
\end{figure}

\noindent\textbf{Memory Usage Comparison.}
We illustrate the effectiveness of SNELL in terms of memory usage by comparing it with various PEFT methods.
Figure~\ref{fig: memory_analysis}(a) shows the accuracy and memory usage of different methods on ViT-B/16. Although some methods achieve satisfactory performance, their memory usage is excessively large, even surpassing that of full fine-tuning~(\textit{e.g.} SPT-Adapter and VPT-Deep).
In comparison, SNELL achieves superior performance on downstream tasks with memory usage comparable to memory-efficient methods, including LoRA and Adapter.

Additionally, we present the memory usage evolutions during the fine-tuning process in Figure~\ref{fig: memory_analysis}(b) to provide a detailed explanation of how SNELL can save memory. 
In the model initialization stage, SNELL exhibits a significantly smaller memory usage compared to full fine-tuning. This is because full fine-tuning stores all weight matrices as learnable parameters in the optimizer, whereas SNELL only stores low-rank matrices with smaller parameter volumes.
In the feed-forward phase, the memory usage increases with the storage of intermediate variables for backpropagation. 
Unlike other intermediate variables, the adaptation matrix $\Delta \mathbf{W}$ in SNELL solely relies on the low-rank parameter matrices, which are already stored in the optimizer. Therefore, it can be dumped in the feed-forward phase and recovered in backpropagation immediately, saving from a large amount of memory usage~(SNELL \textit{vs.} SNELL storing $\Delta \mathbf{W}$).

\noindent\textbf{Scaling to Larger Models.} 
To investigate the scalability of SNELL to large models, we apply it to ViT models of varying sizes~(ViT-B/16, ViT-L/16, and ViT-H/16 pre-trained on ImageNet21K).
We follow the experimental setup presented in Section~\ref{sec: implementation}, except for modifying the batch size for experiments on ViT-H/14 to 8 and changing the search scope of sparsity ratios $s\in\{0, 0.9\}$. 

As depicted in Figure~\ref{fig: memory_analysis}(c), the memory usage of full fine-tuning increases rapidly as the model size grows. 
This observation highlights that existing PEFT methods like VPT and SPT, despite their advanced performances, incur substantial memory costs when applied to large-scale models due to even higher memory usage than full fine-tuning.
In contrast, SNELL exhibits a notable advantage in terms of memory usage for larger models~(similar to LoRA-8). When applied to ViT-H/14, the memory usage of SNELL is only approximately 50\% of that required for full fine-tuning, exemplifying its significant memory-saving capability on large models.

Regarding the performance, as shown in Table~\ref{tab: app_performance}, SNELL-8 outperforms LoRA-8 on all dataset groups~(Natural, Specialized, and Structured) as well as the mean accuracy for both ViT-L and ViT-H on the VTAB-1k benchmark.
This demonstrates the effectiveness of SNELL for adapting large pre-trained models to downstream tasks.

\begin{table}[]
\centering
\footnotesize
\caption{Comparisons on VTAB-1k benchmark with supervised pre-trained ViT-L/16 and ViT-H/16. Top-1 accuracy is reported. The best result is in \textbf{bold}.}
\setlength\tabcolsep{4.5pt}
\begin{tabular}{@{}ccccc|cccc@{}}
\toprule
\multirow{2}{*}{Methods} & \multicolumn{4}{c|}{VTAB-1k ViT-L/16}                  & \multicolumn{4}{c}{VTAB-1k ViT-H/14}                   \\ \cmidrule(l){2-9} 
                         & Natural & Specialized & Structured & Mean Acc. & Natural & Specialized & Structured & Mean Acc. \\ \midrule
LoRA-8                   & 81.2    & 86.6        & 53.4       & 73.7      & 77.9    & 84.8        & 55.9       & 72.9      \\
SNELL-8                  & 82.3    & 86.9        & 56.6       & \textbf{75.3}      & 79.5    & 85.1           & 56.9          & \textbf{73.8}         \\ \bottomrule
\end{tabular}
\vspace{-0.4cm}
\label{tab: app_performance}
\end{table}

\subsection{Ablation Studies}

\noindent\textbf{Effect of Kernelized LoRA.}
We explore the effectiveness of kernelized LoRA by comparing the performance of sparsifying a full-rank matrix, the merged adaptation matrix of LoRA, and the merged adaptation matrix of kernelized LoRA.  
Experimental results are presented in Table~\ref{tab: ablation}(a).
We can see that sparsifying the merged adaptation matrix of LoRA significantly underperforms a full-rank matrix.
This reveals that the low-rank property of the merged adaptation matrix in LoRA greatly compromises the weight selection scope, leading to performance degradation for sparse tuning.
However, when we replace LoRA with kernelized LoRA, the performance becomes notably comparable to that of the full-rank matrix under the strong sparsity constraint ($s=0.9$).
This indicates that kernelized LoRA can effectively leverage sparse tuning while maintaining a low memory usage.
\begin{table}[t]
	\centering
    \caption{
    (a) Performance on VTAB-1k of sparsifying a full-rank matrix, the merged adaptation matrix of LoRA-8 and kernelized LoRA-8~(KLoRA-8) with sparsity ratio $s=0.9$. 
    (b) The mean accuracy on VTAB-1k of kernelized LoRA~(KLoRA) and SNELL~(KLoRA+sparsifying) with different ranks of learnable matrices. Perf. Imp. denotes the performance improvement of SNELL over KLoRA.}
	\subfloat[\label{tab: klora_abs}]{
		\scalebox{0.865}{
          \begin{tabular}{@{}ccccc@{}}
            \toprule
            Matrix    & Natural & Specialized & Structured & Mean Acc. \\ \midrule
            Full-Rank    &    80.5     &          85.1   &      57.6      & 74.4           \\ \midrule
            LoRA-8 &    61.1     &  81.2           &  54.7          &  65.7         \\
            KLoRA-8   & 79.4    & 84.5        & 57.9       & 73.9      \\ \bottomrule
            \end{tabular}}
	}
	\subfloat[\label{table: location}]{
		\scalebox{0.865}{
    \begin{tabular}{@{}cccc@{}}
    \toprule
    Method     & $r=8$  & $r=16$ & $r=32$ \\ \midrule
    KLoRA     & 73.2 & 73.0 & 72.7 \\
    SNELL      & 74.2 & 74.4 & 74.6 \\ \midrule
    Perf. Imp. & +1.0 & +1.4 & +1.9 \\ \bottomrule
    \end{tabular}}
	}
	\label{tab: ablation}
 \vspace{-0.6cm}
\end{table}

\noindent\textbf{Effect of Sparse Tuning.}
Table~\ref{tab: ablation}(b) shows the performance comparison between SNELL and kernelized LoRA to explore the effectiveness of sparse tuning.
For kernelized LoRA with different ranks, applying sparse tuning can consistently improve their performance.
Moreover, as the rank of the learnable matrix increases, the performance of kernelized LoRA decreases while that of SNELL increases. 
This difference stems from the model regularization.
Similar to sparse regularization, the low-rank property of LoRA that constrains the dependence between individual weights, can also be taken as a form of regularization.
As the rank of the learnable matrix increases, the effect of low-rank regularization diminishes. Consequently, kernelized LoRA becomes more susceptible to over-fitting and encounters performance degradation.
In contrast, SNELL employs both low-rank and sparse regularization.
Higher ranks enable better sparsification towards downstream tasks, boosting sparse regularization that counteracts the diminished low-rank regularization.
Therefore, a higher rank may lead to over-fitting in kernelized LoRA, but it can further enhance performance with sparse tuning.

\begin{figure}
  \centering
  \includegraphics[width=0.95\linewidth]{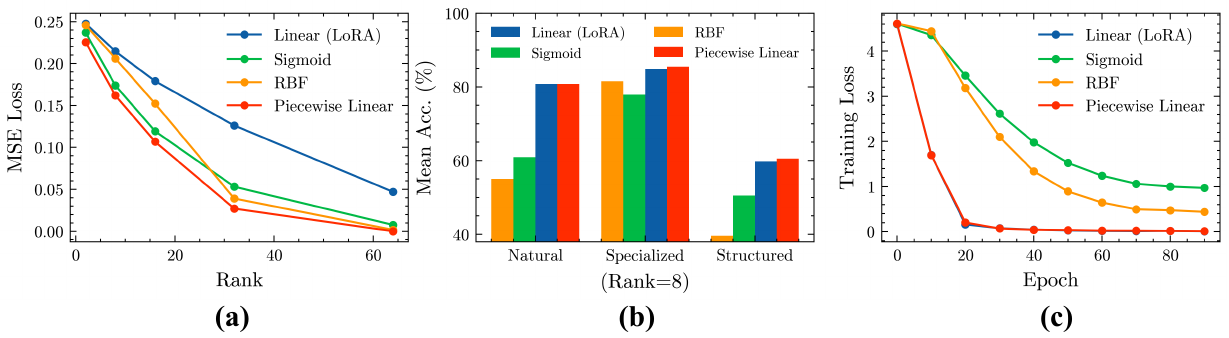}
  \vspace{-0.1cm}
  \caption{(a) The fitting ability of different kernel functions.
  We fit random sparse matrices by merging two learnable low-rank matrices with different kernel functions and compute the MSE loss.
  (b) Performance comparison on groups of datasets in VTAB-1k. (c) Training loss on CIFAR-100 dataset in VTAB-1k benchmark of kernelized LoRA with different kernel functions.}
  \label{fig: kernel_func}
  \vspace{-0.1cm}
\end{figure}

\noindent\textbf{Effect of Different Kernel Function.}
We investigate the effectiveness of different kernel functions in kernelized LoRA. First, we explore the ability of different kernel functions to fit randomly generated full-rank matrices based on low-rank matrices using the gradient descent algorithm~(introduced in Appendix~\ref{sec: app_details_fig4}). As shown in Figure~\ref{fig: kernel_func}(a), we explored four kinds of kernel functions. Compared with the linear kernel function, nonlinear kernel functions can reconstruct the full-rank target matrix more accurately based on the low-rank matrices.
Subsequently, we explore the performance of kernelized LoRA with different kernel functions for pre-trained model fine-tuning in Figure~\ref{fig: kernel_func}(b).
We find that using piecewise linear distance as the kernel function can achieve better results compared to linear kernel function~(LoRA), while using Sigmoid and RBF kernels leads to severe performance degradation. This is because the complex non-linear kernel functions such as the exponential function increase the optimization difficulty in deep networks shown in Figure~\ref{fig: kernel_func}(c). 
More comparisons between LoRA and kernelized LoRA~(with piecewise linear kernel) are presented in Appendix~\ref{sec: app_klora_abl}.

\begin{figure}
  \centering
  \includegraphics[width=0.95\linewidth]{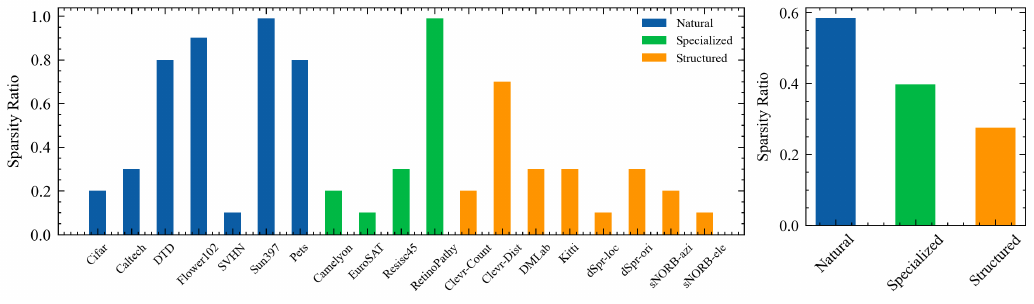}
  \vspace{-0.2cm}
  \caption{The optimal sparsity ratio of SNELL-8 on different downstream tasks~(\textit{left}) and the average optimal sparsity ratio within each group~(\textit{right}) in VTAB-1k benchmark. The pre-trained model is the ConvNeXt-B pre-trained on ImageNet-21k.}
  \label{fig: optimal_ratio}
  \vspace{-0.5cm}
\end{figure}

\noindent\textbf{Optimal Sparsity Ratio for Different Downstream Tasks.}
We provide the optimal sparsity ratio of SNELL-8 on tasks from VTAB-1k benchmarks in Figure~\ref{fig: optimal_ratio}.
The optimal sparsity ratio varies significantly across different downstream tasks within the same group~(\textit{e.g.}, Cifar \textit{vs.} Sun397, dSpr-loc \textit{vs.} Clevr-Dist).
Furthermore, we can observe that the Natural task group exhibits a higher average optimal sparsity ratio compared to the Specialized group, while the Structured group demonstrates the lowest ratio.
This observation aligns with the example illustrated in Figure~\ref{fig: dataset_vis}, where cross-domain adaptation from a model pre-trained on natural images~(ImageNet) to images of Specialized and Structured groups necessitates a larger number of tunable parameters.

\section{Conclusion}
In this work, we proposed a PEFT method named SNELL~(\textbf{S}parse tuning with ker\textbf{NEL}ized \textbf{L}oRA) to conduct high-performance sparse tuning with low memory usage.
To reduce memory usage, we sparsified the adaptation matrix merged with low-rank matrices rather than the pre-trained weight matrix to reduce the volume of learnable parameters stored in the optimizer. Then we designed a competition-based sparsification mechanism to avoid the additional memory usage of storing the tunable weight indexes.
To reveal the effectiveness of sparse tuning, we utilize nonlinear kernel functions to merge the adaptation matrix, increasing the rank of the merged matrix to maintain a compact representation suitable for sparse tuning with low memory usage.
Extensive experiments demonstrated the ability of SNELL to leverage the high performance of sparse tuning and the low memory usage of LoRA.
For future work, we will apply SNELL on larger models such as LLMs and improve its training efficiency.
For limitations discussion, please refer to Appendix~\ref{sec: app_limitation}.

\section*{Acknowledgement}
This work was supported in part by the National Key R$\&$D Program of China under Grant 2023YFC2508704, in part by the National Natural Science Foundation of China: 62236008, U21B2038, and 61931008, and in part by the Fundamental Research Funds for the Central Universities.
The authors would like to thank Zhengqi Pei, Yue Wu, and the anonymous reviewers for their constructive comments and suggestions that improved this manuscript.

\clearpage

{
\bibliographystyle{plain}
\bibliography{neurips_2024}
}
\newpage
\appendix

\renewcommand{\thefigure}{A\arabic{figure}}
\renewcommand{\thetable}{A\arabic{table}}
\renewcommand{\theequation}{A\arabic{equation}}

\section{More Details of Experimental Setup}
\label{sec: app_details}
\subsection{More Details of Contenders.}
\label{sec: app_contenders}
\begin{itemize}[leftmargin=*]
    \item Full: fully tunes all the model parameters~(including backbone and classification head).
    \item Linear: freezes all the backbone parameters and only tunes the linear classification head.
    \item Bias~\cite{zaken2021bitfit}: freezes all model parameters except for the bias term and the linear classification head.
    \item Partial-1: freezes the backbone except for the last 1 layer and also tunes the classification head as described in \cite{jia2022visual}.
    \item MLP-3: freezes the backbone and tunes the classification head implemented by a trainable 3-layer multi-layer percepton as described in \cite{jia2022visual}.
    \item VPT-Shallow~\cite{jia2022visual}: freezes all the backbone parameters while introducing additional trainable prompts to the input space of the pretrained ViT.
    \item VPT-Deep~\cite{jia2022visual}: freezes the backbone while appending additional trainable prompts to the sequence in the multi-head self-attention layer of each ViT block.
    \item Adapter-$r$~\cite{houlsby2019parameter}: freezes all the backbone parameters while adding a down projection, a ReLU non-linearity, and an up projection layer sequentially in the feed-forward network (FFN) of each visual Transformer block. We report the performance implemented by \cite{he2023sensitivity} for comparison.
    \item Lora-$r$~\cite{hu2021lora}: freezes all the backbone parameters while adding a concurrent branch including two low-rank matrices to the weight matrices in the multi-head self-attention layers to approximate efficiently updating them. The low-rank matrices can be merged into the backbone weights after fine-tuning. We report the performance implemented by \cite{he2023sensitivity} for comparison.
    \item SPT~\cite{he2023sensitivity}: identifies the tunable parameters for a given task in a data-dependent way, and utilizes LoRA~(SPT-LoRA) or Adapter~(SPT-Adapter) for weight matrices with a large number of tunable parameters and sparse tuning for weight matrices with a small number of tunable parameters. 
    \item VQT~\cite{tu2023visual}: introduces a handful of learnable query tokens to each layer for adaptation.
    \item DoRA~\cite{liu2024dora}: decomposes the pre-trained weight into two components, {\it i.e.}, magnitude and direction, for fine-tuning. It specifically employs LoRA for directional updates to efficiently minimize the number of trainable parameters.
    \item GPS~\cite{DBLP:conf/cvpr/ZhangZGZSZZ24}: identifies task-dependent tunable weights and applies sparse tuning to these weights.
\end{itemize}

\subsection{Dataset Samples for the Downstream Tasks}
\begin{figure}
  \centering
  \includegraphics[width=1.0\linewidth]{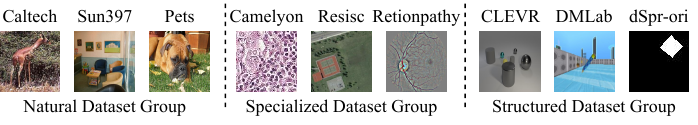}
  \caption{Samples of different downstream tasks in VTAB-1k.}
  \label{fig: dataset_vis}
  \vspace{-0.3cm}
\end{figure}

We visualize some sampled images from different downstream tasks of VTAB-1k~\cite{zhai2019large}) in Figure~\ref{fig: dataset_vis}. 
The VTAB-1k benchmark encompasses a diverse range of tasks, including natural images, remote sensing, medical images, etc. Notably, our SNELL has achieved state-of-the-art (SOTA) performance on these datasets, demonstrating its general effectiveness.

\subsection{Implementation details of Figure~\ref{fig: kernel_func}(a)}
\label{sec: app_details_fig4}
Given a random matrix $\mathbf{W}^{(gt)}\in\mathbb{R}^{m\times n}$, we fit this matrix by merging two low-rank learnable matrices $\mathbf{B}\in\mathbb{R}^{m\times r},\mathbf{A}\in\mathbb{R}^{n\times r}$ with different kernel functions $\kappa$,
\begin{equation}
    \min\limits_{\mathbf{A},\mathbf{B}} \frac{1}{mn}\sum\limits_{i=1}^m\sum\limits_{j=1}^n(\mathbf{W}^{(gt)}_{ij}-\kappa(\mathbf{B}_{i,\cdot},\mathbf{A}_{j,\cdot}))^2.
\end{equation}
We use gradient descent for $1e5$ optimization steps, employing the Adam optimizer with a learning rate of $1e-4$.
We fit 10 randomly generated matrices for each kernel function presented in Table~\ref{tab: app_kernel} and report the average MSE Loss in Figure~\ref{fig: kernel_func}(a).

\section{Introduction of Utilized Kernel Functions}
\label{sec: app_kernel}
\noindent\textbf{Kernel Function Definition~(positive semi-definite).}
Consider a vector space $\mathbb{R}^r$, a kernel function $\kappa: \mathbb{R}^r\times \mathbb{R}^r \rightarrow \mathbb{R}$ is called a positive semi-definite kernel on $\mathbb{R}^r$ if 
\begin{equation}
    \sum\limits_{i=1}^n\sum\limits_{j=1}^n c_ic_j\kappa(\mathbf{x}_i, 
    \mathbf{x}_j)\ge 0
\end{equation}
holds for all $\mathbf{x}_1, ..., \mathbf{x}_n \in \mathbb{R}^r, c_1, ..., c_n \in \mathbb{R}, n\in\mathbb{N}$.

Given two vectors $\mathbf{x}, \mathbf{x'}\in\mathbb{R}^{r}$, we show the utilized kernel functions in Table~\ref{tab: app_kernel}.
We introduce additional learnable parameters~(the \textit{e.g.} $\alpha$ for Sigmoid and RBF kernel, $\alpha_p$ for piecewise linear kernel) that enable the merged adaptation matrix $\Delta \mathbf{W}$ to accommodate both positive and negative values. The additional parameters select certain elements in the matrix and assign them negative values, without compromising the high-rank property of the merged adaptation matrix $\Delta \mathbf{W}$.
We set $P=2$ for the piecewise linear kernel.
\begin{table}[]
\centering
\caption{Expression of kernel function utilized in the main text.}
\begin{tabular}{@{}ll@{}}
\toprule
Kernel Function  & Expression \\ \midrule
Linear           & $\kappa(\mathbf{x}, \mathbf{x'})=\mathbf{x}^\top \mathbf{x'}$           \\
Piecewise Linear & $\kappa(\mathbf{x}, \mathbf{x'})=\sum\limits_{p=1}^P\alpha_p\Vert \mathbf{x}_{\lceil rp/P\rceil :\lceil r(p+1)/P\rceil}-\mathbf{x '}_{\lceil rp/P\rceil :\lceil r(p+1)/P\rceil} \Vert_2$           \\
Sigmoid          & $\kappa(\mathbf{x}, \mathbf{x'})=\alpha (1+exp(-\beta\mathbf{x}^\top \mathbf{x'}))^{-1} + \gamma$          \\
RBF              & $\kappa(\mathbf{x}, \mathbf{x'})=\alpha (exp(-\beta \Vert \mathbf{x}-\mathbf{x'}\Vert_2^2))) + \gamma$           \\ \bottomrule
\end{tabular}
\label{tab: app_kernel}
\end{table}

\begin{table}[]
\centering
\caption{Top-1 accuracy (\%) on VTAB-1k benchmarks using ViT-B/16 backbone pre-trained on ImageNet-21k supervisedly. The best result is in \textbf{bold}, and the second-best result is {\ul underlined}.}
\setlength\tabcolsep{2.8pt}
\scalebox{0.75}{
\begin{tabular}{@{}ccccccccccccccccccccc@{}}
\toprule
\multicolumn{1}{c|}{\multirow{2}{*}{Methods}} & \multicolumn{7}{c|}{Natural}                                                                                                       & \multicolumn{4}{c|}{Specialized}                                                   & \multicolumn{8}{c|}{Structured}                                                                                                                    & -       \\ \cmidrule(l){2-21} 
\multicolumn{1}{c|}{}                         & \rotatebox{90}{Cifar100}      & \rotatebox{90}{Caltech101}    & \rotatebox{90}{DTD}           & \rotatebox{90}{Flower102}     & \rotatebox{90}{SVHN}          & \rotatebox{90}{Sun397}        & \multicolumn{1}{c|}{\rotatebox{90}{Pets}}          & \rotatebox{90}{Camelyon}      & \rotatebox{90}{EuroSAT}       & \rotatebox{90}{Resisc45}      & \multicolumn{1}{c|}{\rotatebox{90}{Retinopathy}}   & \rotatebox{90}{Clevr-Count}   & \rotatebox{90}{Clevr-Dist}    & \rotatebox{90}{DMLab}         & \rotatebox{90}{KITTI-Dist}    & \rotatebox{90}{dSpr-Loc}      & \rotatebox{90}{dSpr-Ori}      & \rotatebox{90}{sNORB-Azim}    & \multicolumn{1}{c|}{\rotatebox{90}{sNORB-Ele}}     & \rotatebox{90}{Mean Acc.}     \\ \midrule
\multicolumn{1}{c|}{Full}                     & 68.9          & 87.7          & 64.3          & 97.2          & 87.4          & 38.8          & \multicolumn{1}{c|}{86.9}          & 79.7          & 95.7          & 84.2          & \multicolumn{1}{c|}{73.9}          & 56.3          & 58.6          & 41.7          & 65.5          & 57.5          & 46.7          & 25.7          & \multicolumn{1}{c|}{29.1}          & 65.6          \\ \midrule
\multicolumn{21}{c}{Additional-based methods}                                                                                                                                                                                                                                                                                                                                                                                                \\ \midrule
\multicolumn{1}{c|}{MLP-3}                    & 63.8          & 84.7          & 62.3          & 97.4          & 32.5          & 49.2          & \multicolumn{1}{c|}{84.7}          & 77.0          & 88.0          & 70.2          & \multicolumn{1}{c|}{56.1}          & 47.8          & 32.8          & 32.3          & 58.1          & 12.9          & 21.2          & 15.2          & \multicolumn{1}{c|}{24.8}          & 53.2          \\
\multicolumn{1}{c|}{VPT-Shallow}              & {\ul 77.7}    & 86.9          & 62.6          & 97.5          & 74.5          & 51.2          & \multicolumn{1}{c|}{87.3}          & 78.2          & 92.0          & 75.6          & \multicolumn{1}{c|}{72.9}          & 50.5          & 58.6          & 40.5          & 67.1          & 68.7          & 36.1          & 20.2          & \multicolumn{1}{c|}{34.1}          & 64.9          \\
\multicolumn{1}{c|}{VPT-Deep}                 & \textbf{78.8} & 90.8          & 65.8          & 98.0          & 78.1          & 49.6          & \multicolumn{1}{c|}{88.3}          & 81.8          & {\ul 96.1}    & 83.4          & \multicolumn{1}{c|}{68.4}          & 68.5          & 60.0          & 46.5          & 72.8          & 73.6          & 47.9          & 32.9          & \multicolumn{1}{c|}{37.8}          & 69.4          \\
\multicolumn{1}{c|}{Adapter-8}                & 69.2          & 90.1          & 68.0          & 98.8          & 82.8          & 54.3          & \multicolumn{1}{c|}{89.9}          & 84.0          & 94.9          & 81.9          & \multicolumn{1}{c|}{75.5}          & 80.9          & 65.3          & 48.6          & 78.3          & 74.8          & 48.5          & 29.9          & \multicolumn{1}{c|}{{\ul 41.6}}    & 71.4          \\
\multicolumn{1}{c|}{Adapter-32}               & 68.7          & 92.2          & 69.8          & 98.9          & 84.2          & 53.0          & \multicolumn{1}{c|}{90.3}          & 83.2          & 95.4          & 83.2          & \multicolumn{1}{c|}{74.3}          & 81.9          & 63.9          & 48.7          & 80.6          & 76.2          & 47.6          & 30.8          & \multicolumn{1}{c|}{36.4}          & 71.5          \\
\multicolumn{1}{c|}{NOAH}                     & 69.6          & 92.7          & 70.2          & 99.1          & 86.1          & 53.7          & \multicolumn{1}{c|}{90.4}          & 84.4          & 95.4          & 83.9          & \multicolumn{1}{c|}{75.8}          & 82.8          & \textbf{68.9} & 49.9          & {\ul 81.7}    & 81.8          & 48.3          & 32.8          & \multicolumn{1}{c|}{\textbf{44.2}} & 73.2          \\
\multicolumn{1}{c|}{SPT-Adapter}              & 72.9          & 93.2          & {\ul 72.5}    & \textbf{99.3} & 88.8          & {\ul 55.8}    & \multicolumn{1}{c|}{{\ul 91.4}}    & \textbf{86.2} & {\ul 96.1}    & 85.5          & \multicolumn{1}{c|}{75.5}          & 83.0          & 68.0          & 51.9          & 81.2          & 82.4          & \textbf{51.9} & 31.7          & \multicolumn{1}{c|}{41.2}          & 74.1          \\ \midrule
\multicolumn{21}{c}{Reparameterized-based methods}                                                                                                                                                                                                                                                                                                                                                                                           \\ \midrule
\multicolumn{1}{c|}{Linear}                   & 63.4          & 85.0          & 63.2          & 97.0          & 36.6          & 51.0          & \multicolumn{1}{c|}{86.3}          & 78.5          & 87.5          & 68.6          & \multicolumn{1}{c|}{74.0}          & 34.3          & 30.6          & 33.2          & 55.4          & 12.5          & 20.0          & 9.6           & \multicolumn{1}{c|}{19.2}          & 52.9          \\
\multicolumn{1}{c|}{Partial-1}                & 66.8          & 85.9          & 62.5          & 97.3          & 37.6          & 50.6          & \multicolumn{1}{c|}{85.5}          & 78.6          & 89.8          & 72.5          & \multicolumn{1}{c|}{73.3}          & 41.5          & 34.3          & 33.9          & 61.0          & 31.3          & 32.8          & 16.3          & \multicolumn{1}{c|}{22.4}          & 56.5          \\
\multicolumn{1}{c|}{Bias}                     & 72.8          & 87.0          & 59.2          & 97.5          & 59.9          & 51.4          & \multicolumn{1}{c|}{85.3}          & 78.7          & 91.6          & 72.9          & \multicolumn{1}{c|}{69.8}          & 61.5          & 55.6          & 32.4          & 55.9          & 66.6          & 40.0          & 15.7          & \multicolumn{1}{c|}{25.1}          & 62.0          \\
\multicolumn{1}{c|}{LoRA-8}                   & 67.1          & 91.4          & 69.4          & 98.8          & 85.3          & 54.0          & \multicolumn{1}{c|}{90.4}          & 84.9          & 95.3          & 84.4          & \multicolumn{1}{c|}{73.6}          & 82.9          & 69.2          & 49.8          & 78.5          & 75.7          & 47.1          & 31.0          & \multicolumn{1}{c|}{44.0}          & 72.3          \\
\multicolumn{1}{c|}{LoRA-16}                  & 68.1          & 91.4          & 69.8          & 99.0          & 86.4          & 53.1          & \multicolumn{1}{c|}{90.5}          & 85.1          & 95.8          & 84.7          & \multicolumn{1}{c|}{74.2}          & 83.0          & 66.9          & 50.4          & 81.4          & 80.2          & 46.6          & 32.2          & \multicolumn{1}{c|}{41.1}          & 72.6          \\
\multicolumn{1}{c|}{SPT-LoRA}                 & 73.5          & {\ul 93.3}    & {\ul 72.5}    & \textbf{99.3} & 87.9          & 55.5          & \multicolumn{1}{c|}{\textbf{91.5}} & {\ul 85.7}    & \textbf{96.2} & 85.9          & \multicolumn{1}{c|}{75.9}          & \textbf{84.4} & 67.6          & 52.5          & \textbf{82.0} & 81.0          & {\ul 51.1}    & 30.2          & \multicolumn{1}{c|}{41.3}          & 74.1          \\ \midrule
\multicolumn{1}{c|}{SNELL-8}                  & 73.7          & 92.7          & 72.4          & {\ul 99.2}    & 89.2          & 55.4          & \multicolumn{1}{c|}{{\ul 91.4}}    & 84.9          & {\ul 96.1}    & {\ul 86.4}    & \multicolumn{1}{c|}{75.2}          & {\ul 84.0}    & 68.5          & \textbf{53.5} & 81.0          & {\ul 82.7}    & 49.9          & {\ul 33.9}    & \multicolumn{1}{c|}{39.2}          & 74.2          \\
\multicolumn{1}{c|}{SNELL-16}                 & 74.2          & \textbf{93.4} & {\ul 72.5}    & \textbf{99.3} & {\ul 90.2}    & 55.7          & \multicolumn{1}{c|}{{\ul 91.4}}    & {\ul 85.7}    & 95.8          & \textbf{86.5} & \multicolumn{1}{c|}{\textbf{76.3}} & \textbf{84.4} & 68.2          & {\ul 53.0}    & \textbf{82.0} & 82.2          & 49.6          & 33.3          & \multicolumn{1}{c|}{40.6}          & {\ul 74.4}    \\
\multicolumn{1}{c|}{SNELL-32}                 & 74.5          & \textbf{93.4} & \textbf{73.1} & \textbf{99.3} & \textbf{91.1} & \textbf{55.9} & \multicolumn{1}{c|}{\textbf{91.5}} & 85.5          & {\ul 96.1}    & \textbf{86.5} & \multicolumn{1}{c|}{{\ul 76.2}}    & 83.4          & {\ul 68.6}    & 52.2          & 81.3          & \textbf{83.2} & 50.7          & \textbf{35.9} & \multicolumn{1}{c|}{39.0}          & \textbf{74.6} \\ \bottomrule
\end{tabular}}
\label{tab: vtab1k_performance}
\end{table}

\section{Additional Experiments}
\subsection{More Comparisons with Existing Methods}
\label{sec: app_more_compare}
Given that some methods do not provide performance or implementation details on both FGVC and VTAB benchmarks, we present a comparison between SNELL and these methods in the appendix rather than in Table~\ref{tab: fgvc_vtab1k}. First, we provide comparisons with GPS~\cite{DBLP:conf/cvpr/ZhangZGZSZZ24} on the FGVC benchmark in terms of performance and memory usage in Table~\ref{tab: gps_compare}. With comparable performance, SNELL has a significant advantage over GPS in terms of memory usage. Then, we compare SNELL with VQT~\cite{tu2023visual} on VTAB-1k dataset in Table~\ref{tab: vqt_compare}. SNELL significantly outperforms VQT~(76.9\% {\it vs.} 68.8\%).

\begin{table}[t]
	\centering
    \caption{
    (a) Performance comparisons on FGVC between SNELL and GPS. 
    (b) Memory usage comparison between SNELL and GPS~(batchsize=8) on different pre-trained models.}
	\subfloat[\label{tab: klora_abs}]{
		\scalebox{0.865}{
          \begin{tabular}{@{}ccccccc@{}}
\toprule
Method   & CUB  & NABirds & \begin{tabular}[c]{@{}c@{}}Oxford\\ Flowers\end{tabular} & \begin{tabular}[c]{@{}c@{}}Stanford\\ Dogs\end{tabular} & \begin{tabular}[c]{@{}c@{}}Stanford\\ Cars\end{tabular} & Mean \\ \midrule
GPS      & {\bf 89.9} & 86.7    & \textbf{99.7}                                                     & \textbf{92.2}                                                    & 90.4                                                    & \textbf{91.8} \\
SNELL-32 & {\bf 89.9} & {\bf 87.0}    & 99.4                                                     & 92.0                                                    & \textbf{90.5}                                                    & \textbf{91.8} \\ \bottomrule
\end{tabular}}
	}
	\subfloat[\label{table: location}]{
		\scalebox{0.81}{
    \begin{tabular}{@{}c|ccc@{}}
\toprule
\multirow{2}{*}{Method} & \multicolumn{3}{c}{Memory Usage (Mb)} \\ \cmidrule(l){2-4} 
                        & ViT-B       & ViT-L      & ViT-H      \\ \midrule
GPS                     & 2428        & 7522       & 16119      \\
SNELL-32                & \textbf{1673}        & \textbf{4519}       & \textbf{9692}       \\ \bottomrule
\end{tabular}}
	}
	\label{tab: gps_compare}
 \vspace{-0.6cm}
\end{table}

\begin{table}[]
\centering
\caption{Performance comparisons on VTAB-1k between SNELL and VQT with ViT-B/16 pre-trained on ImageNet-21K. The best result is in {\bf bold}. }
\begin{tabular}{@{}lcccc@{}}
\toprule
Method & Natural & Specialized & Structured & Mean Acc. \\ \midrule
VQT             & 72.7             & 84.5                 & 49.3                & 68.8               \\
SNELL-8         & 82.0             & 85.7                 & 61.6                & 76.4               \\
SNELL-16        & 82.4             & \textbf{86.1}        & 61.7                & 76.7               \\
SNELL-32        & \textbf{82.7}    & \textbf{86.1}        & \textbf{61.8}       & \textbf{76.9}      \\ \bottomrule
\end{tabular}
\label{tab: vqt_compare}
\end{table}

\subsection{Per-task results on the VTAB-1k benchmark}
We provide the per-tasks results on the VTAB-1k benchmark using ViT-B/16 supervised pre-trained on ImageNet21K in Table~\ref{tab: vtab1k_performance}.
Our SNELL has demonstrated superior performance by achieving SOTA performance on 13 downstream tasks. Additionally, SNELL achieves SOTA performance on the mean accuracy across all tasks~(74.6\% \textit{vs.} 74.1\%), indicating its effectiveness in various domains.

\subsection{More ablation studies.}
\subsubsection{Comparison between Competition-based Sparsification and Pre-defined Weight Mask}
To verify the effectiveness of the proposed competition-based sparsification mechanism, we compare the performance on FGVC datasets between kernelized LoRA (KLoRA-8-Fixed) with pre-defined fixed masks, and SNELL in Table~\ref{tab: fixed_mask}. The weight masks are generated by SPT~\cite{he2023sensitivity}. For a fair comparison, we utilize the same data augmentation as SPT. Compared to our dynamic masking strategy, pre-defined fixed masking can hardly identify and adjust the most task-relevant weights in an end-to-end fashion, which leads to performance degradation (89.4 vs. 90.3).
\begin{table}[]
\small
\centering
\caption{Performance comparisons on FGVC benchmark between kernelized LoRA with fixed weight masks~(KLoRA-8-Fixed) and our dynamical masks~(SNELL-8). }
\begin{tabular}{@{}lcccccc@{}}
\toprule
Method        & CUB-200 & NABirds & Oxford Flowers & Stanford Dogs & Stanford Cars & Mean \\ \midrule
KLoRA-8-Fixed & 88.0    & 82.1    & 99.0           & 89.4          & 88.4          & 89.4 \\
SNELL-8       & 89.0    & 83.9    & 99.3           & 90.6          & 88.6          & {\bf 90.3} \\ \bottomrule
\end{tabular}
\label{tab: fixed_mask}
\end{table}

\subsubsection{Comparison between Kernelized LoRA and LoRA}
\label{sec: app_klora_abl}
We compare the performance of LoRA and kernelized LoRA on the VTAB-1k benchmark, where all weight matrices of the pre-trained models are fine-tuned to ensure a fair comparison. The experimental results are presented in Table~\ref{tab: app_klora}.
Through experiments with different ranks, we observed that kernelized consistently outperforms LoRA across various task groups. 
The replacement of the inner product with nonlinear kernel functions leads to stronger expressive ability, which in turn contributes to improved performance on downstream tasks.
\begin{table}[]
\centering
\small
\caption{Comparisons between LoRA and kernelized LoRA~(KLoRA) on VTAB-1k using ViT-B/16 pre-trained on ImageNet21k supervisedly. Better performance for the same rank is in \textbf{bold}.}
\begin{tabular}{@{}ccccc@{}}
\toprule
Method   & Natural       & Specialized   & Structured    & Mean Acc.     \\ \midrule
LoRA-8   & \textbf{80.8} & 84.9          & 59.6          & 75.1          \\
KLoRA-8  & \textbf{80.8} & \textbf{85.5} & \textbf{60.5} & \textbf{75.6} \\ \midrule
LoRA-16  & 80.6          & 85.6          & 58.5          & 74.9          \\
KLoRA-16 & \textbf{80.9} & \textbf{85.7} & \textbf{59.7} & \textbf{75.4} \\ \midrule
LoRA-32  & 79.4          & \textbf{85.4} & 57.8          & 74.2          \\
KLoRA-32 & \textbf{80.8} & \textbf{85.4} & \textbf{59.4} & \textbf{75.2} \\ \bottomrule
\end{tabular}
\label{tab: app_klora}
\end{table}
\subsubsection{Additional Memory Usage from Nonlinear Kernel Functions}
In Figure~\ref{fig: memory_analysis}(a), we observe that SNELL requires additional memory usage compared to LoRA due to the incorporation of nonlinear kernel functions. To explore whether the impact of this additional usage hinders the usability of SNELL on large models, 
we compare the memory usage between SNELL and LoRA on models as the model size grows~(in Table~\ref{tab: memory_comp_lora}). As the model size expands, the incremental memory usage of SNELL becomes negligible.
\begin{table}
	\centering
 \caption{Memory usage comparison between SNELL and LoRA. $\Delta$ Mem. denotes the incremental memory usage of SNELL in comparison to LoRA.}
	\setlength{\tabcolsep}{3pt}
	\scalebox{1.0}{
		\begin{tabular}{@{}lccc@{}}
\toprule
\begin{tabular}[c]{@{}c@{}}{Pre-trained}\\Model\end{tabular} & \begin{tabular}[c]{@{}c@{}}LoRA-8 \\ Mem.~(MB)\end{tabular} & \begin{tabular}[c]{@{}c@{}}SNELL-8 \\ Mem.~(MB)\end{tabular} & \begin{tabular}[c]{@{}c@{}}$\Delta$ Mem. / \\ SNELL-8 Mem. \end{tabular} \\ \midrule
ViT-B/16          & 1546                                                     & 1673                                                      & 0.076                                                              \\
ViT-L/16          & 4325                                                     & 4519                                                      & 0.043                                                              \\
ViT-H/16          & 9325                                                     & 9692                                                      & 0.038                                                              \\ \bottomrule
\end{tabular}}
	
	\label{tab: memory_comp_lora} 
	\vspace{-0.2cm}
\end{table}

\subsection{Experiments on Large Language Models}
We apply SNELL on LLaMA2-7B~\cite{touvron2023llama} to adapt to the commonsense reasoning benchmark. As Table~\ref{tab: llama_result} shows, SNELL achieves a better performance than LoRA. This shows the applicability of SNELL to NLP tasks. Many other vision PEFT approaches lack this capability, as they necessitate a full level of memory usage for fine-tuning as Figure~\ref{fig: memory_analysis}(a) shows. 
\begin{table}[]
\centering
\small
\setlength{\tabcolsep}{4pt}
\caption{Performance on commonsense reasoning benchmark with LLaMA-2-7B.}
\begin{tabular}{@{}lccccccccc@{}}
\toprule
Model    & BoolQ & PIQA & SIQA & HellaSwag & WinoGrande & ARC-e & ARC-c & OBQA & Average       \\ \midrule
LoRA-32  & 69.8  & 79.9 & 79.5 & 83.6      & 82.6       & 79.8  & 64.7  & 81.0 & 77.6          \\
SNELL-32 & 71.4  & 82.9 & 80.7 & 82.1      & 80.9       & 82.6  & 68.0  & 80.8 & \textbf{78.7} \\ \bottomrule
\end{tabular}
\label{tab: llama_result}
\end{table}

\subsection{Training Time Analysis}
\label{sec: app_time}
Table~\ref{tab: app_time} provides a comparison of training time costs between SNELL and other PEFT methods using NVIDIA GeForce RTX 4090 GPU.
The training time of SNELL-8 is slightly higher than LoRA-8~(0.557 \textit{vs.} 0.443). 
By further comparing the training time of SNELL-8 and SNELL-8~(saving $\Delta \mathbf{W}$), it becomes apparent that the increase in time cost primarily stems from the recomputation of the merged adaptation matrix $\Delta \mathbf{W}$. Conversely, the time cost associated with sparsification (KLoRA-8 \textit{vs.} SNELL-8) and kernelization (LoRA-8 \textit{vs.} KLoRA-8~(saving $\Delta \mathbf{W}$)) is relatively small.
Indeed, despite the slight increase in time cost due to the recomputation, one significant advantage is the significant performance improvement and memory efficiency shown in Figure~\ref{fig: memory_analysis}. 
\begin{table}[]
\centering
\caption{Training time cost on ViT-B/16 of different PEFT methods.}
\begin{tabular}{@{}c|c|cc|cc@{}}
\toprule
Method                & LoRA-8 & KLoRA-8 & \begin{tabular}[c]{@{}c@{}}KLoRA-8\\ (saveing $\Delta \mathbf{W}$)\end{tabular} & SNELL-8 & \begin{tabular}[c]{@{}c@{}}SNELL-8\\ (saving $\Delta \mathbf{W}$)\end{tabular} \\ \midrule
Training time (s/img) & 0.443  & 0.522    & 0.446                                                         & 0.557   & 0.448                                                        \\ \bottomrule
\end{tabular}
\label{tab: app_time}
\end{table}

\section{Additional Visualization}

\subsection{Performance of Different Sparsity Ratio}
\begin{figure}
  \centering
  \includegraphics[width=1.0\linewidth]{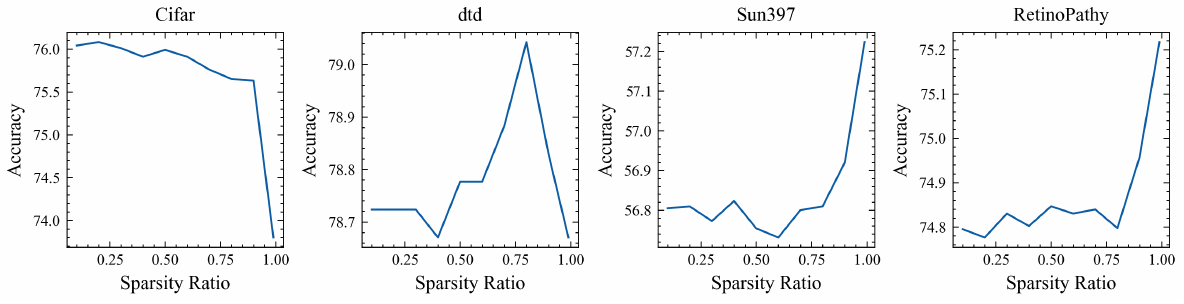}
  \caption{Accuracy of different sparsity ratios using SNELL-8. The pre-trained model is the ConvNeXt-B pre-trained on ImageNet-21k.}
  \label{fig: different_ratio}
  \vspace{-0.3cm}
\end{figure}

Figure~\ref{fig: different_ratio} depicts the accuracy of different sparsity ratios on datasets in VTAB-1k.
Different downstream tasks exhibit diverse preferences for the sparsity ratio. For instance, CIFAR-100 tends to favor a smaller sparsity ratio~(0.2), while DTD prefers a larger sparsity ratio~(0.8). Both Sun397 and Retinopathy tasks also lean towards a larger sparsity ratio~(0.99). 
This highlights the need to consider the specific characteristics of each task when determining the optimal sparsity ratio.

\subsection{Analysis of Tunable Parameters}
\begin{figure}
  \centering
  \includegraphics[width=1.0\linewidth]{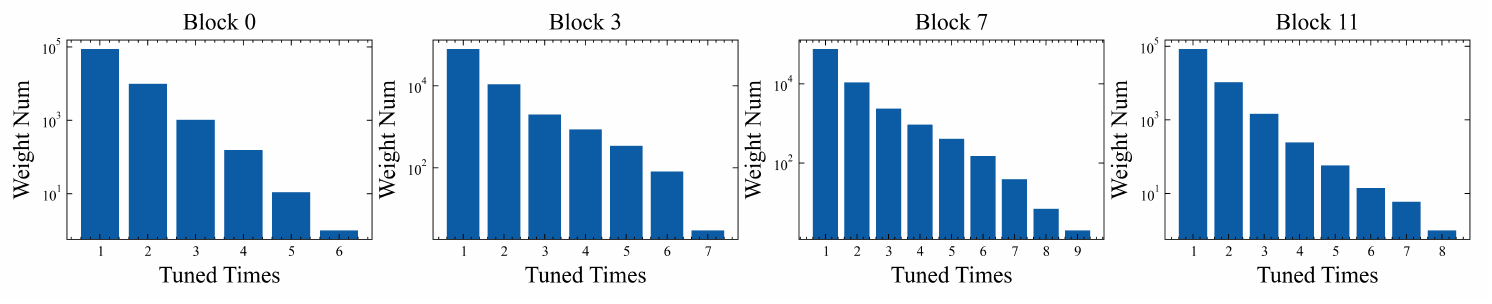}
  \caption{Number of weights in $\mathbf{W}_Q$ selected by multiple tasks with sparsity ratio $s=0.99$.}
  \label{fig: visualize}
  \vspace{-0.3cm}
\end{figure}

We analyze the tunable weights of SNELL for different downstream tasks.
In Figure~\ref{fig: visualize}, we compute the number of weights in the weight matrix $\mathbf{W}_Q$ of self-attention~\cite{vaswani2017attention} selected by multiple tasks.
We find that most of the weights are only selected by a single downstream task (Tuned Times=1).
Moreover, we find that in blocks of different depths, there will be a small part of weights that are selected by multiple downstream tasks, which indicates that there exists a small number of crucial parameters to improve the model's performance on downstream tasks.

\section{Discussion}
\subsection{Tunable Parameter Volume Computing}
We justify our choice to omit to report the volume of learnable parameters.
First, computing the volume of tunable parameters in SNELL is difficult.
In the case of LoRA, the volume corresponds to the shape of the learnable low-rank matrices. Conversely, for sparse tuning, the volume is determined by the number of updated weights.
However, SNELL employs low-rank matrices as learnable parameters and achieves additional updated weight reduction by sparsifying the merged matrices.
When using the parameter volume computation method of LoRA, calculating the reduction in parameters due to sparsification becomes challenging.
Conversely, applying the parameter volume computation method of sparse tuning would be inherently unfair, given that SNELL is specifically optimized using low-rank matrices.
Second, the parameter efficiency is a pathway to achieve high performance and low memory usage rather than an objective for model improvement, because performance improvement and memory usage reduction hold practical value.
In our experiments, SNELL has demonstrated its advantages in terms of high performance and low memory usage, which we consider more valuable than the pursuit of fewer learnable parameters.
\subsection{Limitation Discussion}
\label{sec: app_limitation}
Despite achieving state-of-the-art performance with low memory usage, SNELL requires more training time than LoRA.
The additional training time cost comes from the recomputation of the merged matrix $\Delta \mathbf{W}$ in the backpropagation process presented in Appendix~\ref{sec: app_time}.
However, it is crucial to note that this limitation can be solved.
Firstly, given the unique characteristics of the kernel matrix, more efficient methods~\cite{NEURIPS2022_e7599c4b} can be employed to calculate the merged adaptation matrix $\Delta \mathbf{W}$.
Secondly, by designing appropriate GPU operators, it is possible to avoid explicitly calculating $\Delta \mathbf{W}$~\cite{pei2023dynamics} during the fine-tuning process like LoRA and reparameterize the learnable low-rank matrices into the pre-trained weight matrices after the fine-tuning process.

\newpage
\section*{NeurIPS Paper Checklist}

\begin{enumerate}

\item {\bf Claims}
    \item[] Question: Do the main claims made in the abstract and introduction accurately reflect the paper's contributions and scope?
    \item[] Answer: \answerYes{} 
    \item[] Justification: The main claims made in the abstract and introduction accurately reflect the paper's contributions and scope.

\item {\bf Limitations}
    \item[] Question: Does the paper discuss the limitations of the work performed by the authors?
    \item[] Answer: \answerYes{} 
    \item[] Justification: We discuss the limitation in Appendix~\ref{sec: app_limitation}

\item {\bf Theory Assumptions and Proofs}
    \item[] Question: For each theoretical result, does the paper provide the full set of assumptions and a complete (and correct) proof?
    \item[] Answer: \answerNA{} 
    \item[] Justification: This paper does not include theoretical results.

    \item {\bf Experimental Result Reproducibility}
    \item[] Question: Does the paper fully disclose all the information needed to reproduce the main experimental results of the paper to the extent that it affects the main claims and/or conclusions of the paper (regardless of whether the code and data are provided or not)?
    \item[] Answer: \answerYes{} 
    \item[] Justification: Please refer to Section~\ref{sec: implementation}, Appendix~\ref{sec: app_details} and the \href{https://github.com/ssfgunner/SNELL}{released codes}.

\item {\bf Open access to data and code}
    \item[] Question: Does the paper provide open access to the data and code, with sufficient instructions to faithfully reproduce the main experimental results, as described in supplemental material?
    \item[] Answer: \answerYes{} 
    \item[] Justification: Please refer to the \href{https://github.com/ssfgunner/SNELL}{released codes}.

\item {\bf Experimental Setting/Details}
    \item[] Question: Does the paper specify all the training and test details (e.g., data splits, hyperparameters, how they were chosen, type of optimizer, etc.) necessary to understand the results?
    \item[] Answer: \answerYes{} 
    \item[] Justification: Please refer to Section~\ref{sec: implementation}.

\item {\bf Experiment Statistical Significance}
    \item[] Question: Does the paper report error bars suitably and correctly defined or other appropriate information about the statistical significance of the experiments?
    \item[] Answer: \answerNo{} 
    \item[] Justification: This paper does not report error bars following the practice of previous studies~\cite{jia2022visual, he2023sensitivity,tu2023visual}.

\item {\bf Experiments Compute Resources}
    \item[] Question: For each experiment, does the paper provide sufficient information on the computer resources (type of compute workers, memory, time of execution) needed to reproduce the experiments?
    \item[] Answer: \answerYes{}
    \item[] Justification: Please refer to Figure~\ref{fig: memory_analysis} and Table~\ref{tab: app_time}.
    
\item {\bf Code Of Ethics}
    \item[] Question: Does the research conducted in the paper conform, in every respect, with the NeurIPS Code of Ethics \url{https://neurips.cc/public/EthicsGuidelines}?
    \item[] Answer: \answerYes{} 
    \item[] Justification: This paper conforms with the NeurIPS Code of Ethics.

\item {\bf Broader Impacts}
    \item[] Question: Does the paper discuss both potential positive societal impacts and negative societal impacts of the work performed?
    \item[] Answer: \answerNA{} 
    \item[] Justification: There is no societal impact of the work performed.
    
\item {\bf Safeguards}
    \item[] Question: Does the paper describe safeguards that have been put in place for responsible release of data or models that have a high risk for misuse (e.g., pretrained language models, image generators, or scraped datasets)?
    \item[] Answer: \answerNA{} 
    \item[] Justification: This paper poses no such risks.

\item {\bf Licenses for existing assets}
    \item[] Question: Are the creators or original owners of assets (e.g., code, data, models), used in the paper, properly credited and are the license and terms of use explicitly mentioned and properly respected?
    \item[] Answer: \answerYes{} 
    \item[] Justification: Please refer to Section~\ref{sec: implementation} and Appendix~\ref{sec: app_details}.

\item {\bf New Assets}
    \item[] Question: Are new assets introduced in the paper well documented and is the documentation provided alongside the assets?
    \item[] Answer: \answerYes{} 
    \item[] Justification: Please refer to the \href{https://github.com/ssfgunner/SNELL}{released codes}.

\item {\bf Crowdsourcing and Research with Human Subjects}
    \item[] Question: For crowdsourcing experiments and research with human subjects, does the paper include the full text of instructions given to participants and screenshots, if applicable, as well as details about compensation (if any)? 
    \item[] Answer: \answerNA{}
    \item[] Justification: The paper does not involve crowdsourcing nor research with human subjects.

\item {\bf Institutional Review Board (IRB) Approvals or Equivalent for Research with Human Subjects}
    \item[] Question: Does the paper describe potential risks incurred by study participants, whether such risks were disclosed to the subjects, and whether Institutional Review Board (IRB) approvals (or an equivalent approval/review based on the requirements of your country or institution) were obtained?
    \item[] Answer: \answerNA{} 
    \item[] Justification: This paper does not involve crowdsourcing nor research with human subjects.

\end{enumerate}

\end{document}